\documentclass[10pt,twocolumn,letterpaper]{article}

\usepackage[pagenumbers]{cvpr}              %

\definecolor{cvprblue}{rgb}{0.21,0.49,0.74}
\usepackage[pagebackref,breaklinks,colorlinks,citecolor=cvprblue]{hyperref}

\usepackage{eccvabbrv}

\usepackage{graphicx}
\usepackage{booktabs}
\usepackage{wrapfig}

\usepackage[accsupp]{axessibility}  %

\usepackage[pagebackref]{hyperref}

\usepackage{orcidlink}

\usepackage[disable,textsize=tiny]{todonotes}

\usepackage{amsmath,amsfonts,bm}
\usepackage{graphicx} %
\usepackage{booktabs}

\usepackage[most]{tcolorbox}
\usepackage{algorithm}
\usepackage{appendix}

\newcommand{\recloss}{\mathcal{L}_{\text{Rec}}}
\newcommand{\reclossratio}{\rho_{\text{Rec}}}
\newcommand{\quantloss}{\mathcal{L}_{\text{Q}}}

\newcommand{\aeloss}{\mathcal{L}_{\mathcal{A}}}

\newcommand{\invloss}{\mathcal{L}_{\text{Inv}}}

\newcommand{\mse}{\text{MSE}}

\newcommand{\Mgeneral}{\mathcal{M}}
\newcommand{\Mone}{\mathcal{M}_1}
\newcommand{\Mtwo}{\mathcal{M}_2}
\newcommand{\Mthree}{\mathcal{M}_3}

\newcommand{\tpr}{TPR@1\%FPR\xspace}

\def\eqref#1{equation~\ref{#1}}

\def\1{\bm{1}}

\def\rvc{{\mathbf{c}}}

\def\rvx{{\mathbf{x}}}

\def\rvz{{\mathbf{z}}}
\def\rvphi{{\mathbf{\phi}}}

\def\gA{{\mathcal{A}}}

\def\gD{{\mathcal{D}}}

\def\gE{{\mathcal{E}}}

\def\gG{{\mathcal{G}}}

\def\gM{{\mathcal{M}}}

\def\gQ{{\mathcal{Q}}}
\def\gQinv{{\mathcal{Q}^{-1}}}

\usepackage{xparse}
\usepackage{xspace}
\usepackage{algorithm}%
\usepackage{tikz}
\usepackage{float}
\usepackage{multirow}
\usepackage{multicol}
\usepackage{colortbl}
\usepackage{array}
\newcommand{\PreserveBackslash}[1]{\let\temp=\\#1\let\\=\temp}
\newcolumntype{C}[1]{>{\PreserveBackslash\centering}p{#1}}
\newcolumntype{R}[1]{>{\PreserveBackslash\raggedleft}p{#1}}
\newcolumntype{L}[1]{>{\PreserveBackslash\raggedright}p{#1}}

\usepackage{microtype}
\usepackage{graphicx}
\usepackage{mathtools}
\usepackage{float}
\usepackage{subcaption}
\usepackage{booktabs} %
\usepackage{amssymb}%
\usepackage{pifont}%

\usepackage{bbold}

\usepackage{amsmath}
\usepackage{xspace}

\setlist[itemize]{leftmargin=*}
\setlist[enumerate]{leftmargin=*}

\newcommand{\DCB}{DCB\@\xspace}

\newcommand{\MGI}{MGI\@\xspace}

\newcommand{\IGM}{IGM\@\xspace}
\newcommand{\IGMs}{IGMs\@\xspace}

\newcommand{\IAR}{IAR\@\xspace}
\newcommand{\IARs}{IARs\@\xspace}

\newcommand{\DMs}{DMs\@\xspace}

\newcommand*{\rej}{{\ooalign{\lower.3ex\hbox{$\sqcup$}\cr\raise.4ex\hbox{$\sqcap$}}}}

\usepackage{stackengine}

\usepackage{xcolor}

\graphicspath{{images/}}

\usepackage{booktabs,arydshln}
\makeatletter
\def\adl@drawiv#1#2#3{%
        \hskip.5\tabcolsep
        \xleaders#3{#2.5\@tempdimb #1{1}#2.5\@tempdimb}%
                #2\z@ plus1fil minus1fil\relax
        \hskip.5\tabcolsep}
\newcommand{\cdashlinelr}[1]{%
  \noalign{\vskip\aboverulesep
           \global\let\@dashdrawstore\adl@draw
           \global\let\adl@draw\adl@drawiv}
  \cdashline{#1}
  \noalign{\global\let\adl@draw\@dashdrawstore
           \vskip\belowrulesep}}
\makeatother

\newcommand{\nlp}[1]{}

\newcolumntype{x}[1]{>{\centering\arraybackslash\hspace{0pt}}p{#1}}

\newcommand{\bihe}[1]{\textcolor{black}{#1}}

\newcommand{\ours}{\texttt{DCB}\xspace}

\newif\ifdraft
\drafttrue

\ifdraft
\usepackage[textsize=tiny]{todonotes}
\newcommand{\janek}[1]{\textcolor{violet}{[JD: #1]}}
\newcommand{\mytodo}[1]{\textcolor{red}{[todo: #1]}}
\newcommand{\mycomment}[1]{\textcolor{red}{[comment: #1]}}
\newcommand{\antoni}[1]{\textcolor{magenta}{AK: #1}}
\newcommand{\adam}[1]{\textcolor{cyan}{[AD: #1]}}
\newcommand{\franzi}[1]{\textcolor{brown}{FB: #1}}

\else
\usepackage[disable]{todonotes}
\newcommand{\janek}[1]{}
\newcommand{\mytodo}[1]{}
\newcommand{\mycomment}[1]{}
\newcommand{\antoni}[1]{}
\newcommand{\adam}[1]{}
\newcommand{\franzi}[1]{}
\fi

\drafttrue

\newcommand{\ourtitle}{MGI: Member vs Generated Inference}

\title{\ourtitle} 
\author{%
  Bihe Zhao, Michel Meintz, Juangui Xu, Franziska Boenisch, Adam Dziedzic \\
  {\normalsize \hspace{0em}\texttt{\{bihe.zhao, michel.meintz, juangui.xu, boenisch, adam.dziedzic\}@cispa.de}}\\
  CISPA Helmholtz Center for Information Security \\  
}

\newif\ifcvpr

\cvprtrue 

\begin{document}

\maketitle

\begin{abstract}

As generative models increasingly produce samples that are indistinguishable from human-created content, it becomes difficult to determine whether a given data point was part of a model’s natural training set or was generated by the model itself, especially when models memorize and reproduce training data. We formalize this challenge as \textit{Member vs Generated Inference} (MGI): given a sample and a target generative model, infer whether the sample is a true training member or a generated output of that model. Focusing on image generation, we show that existing membership inference methods systematically misclassify generated samples as training members, while attribution-based methods often misclassify true members as generated. This failure arises because both approaches rely on likelihood-related signals that are similarly elevated for training examples and for the model’s own outputs. To address MGI, we propose \textit{Data Circuit Breaker} (\DCB), a three-stage method that combines complementary signals from a generative model’s autoencoder and latent generator to distinguish training members from generated samples. Across multiple generative models, including image autoregressive and diffusion models, \DCB consistently addresses the shortcomings of membership inference and attribution methods, remains effective even when models reproduce near-duplicates of training samples, and generalizes to challenging \textit{model derivative} settings in which new models are trained on generated data.

\end{abstract}

\section{Introduction}
\label{sec:intro}

\begin{figure}[t]
    \centering
    \includegraphics[width=\linewidth,trim={0 0 2.2cm 0},clip]{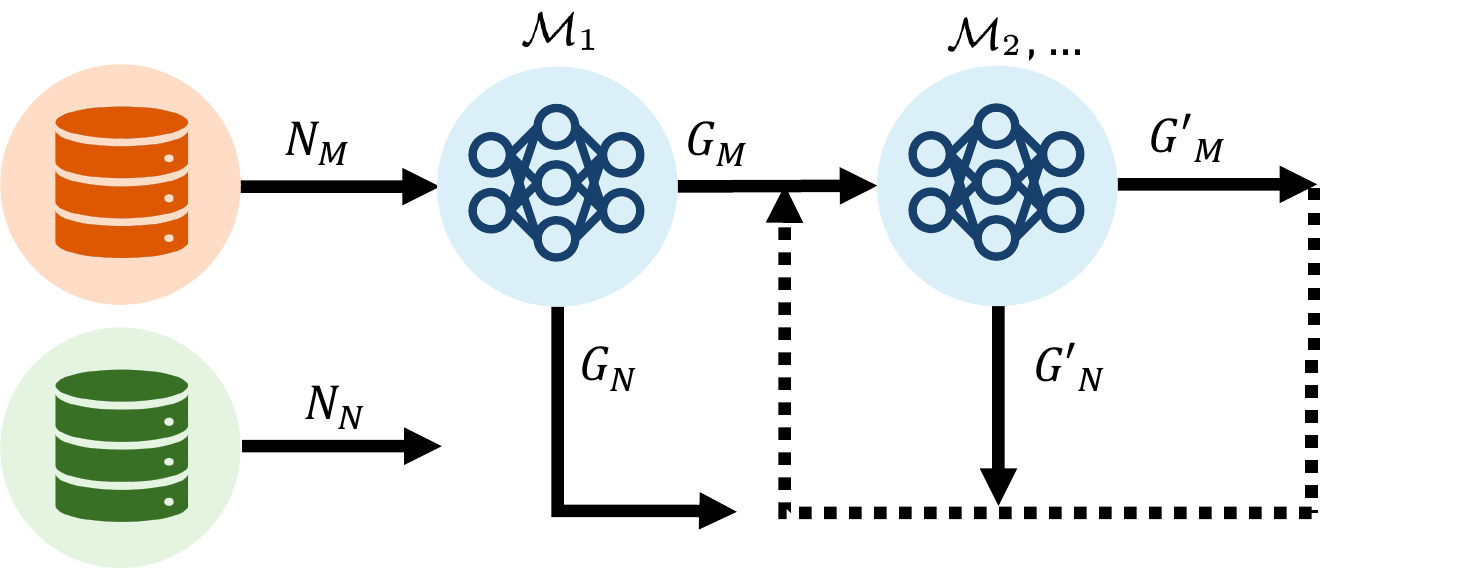}
    \vspace{-10pt}
    \caption{{\normalsize \textbf{Overview of the new Member vs Generated Inference (\MGI) task.} 
     The core challenge is separating genuine training membership from model generation, even across chains of models trained on generated data.
    Let $N = N_M \cup N_N$ denote a natural dataset, where $N_M \cap N_N = \varnothing$. A generative model $\Mone$ is trained on the member set $N_M$, while $N_N$ is held out as natural non-member data. After training, $\Mone$ produces a generated dataset $G = G_M \cup G_N$, with $G_M \cap G_N = \varnothing$. Here, $G_M$ and $G_N$ are both generated by $\Mone$ and therefore follow the same generated-data distribution, but they play different roles in downstream settings: $G_M$ is used to train a new model $\Mtwo$, whereas $G_N$ is withheld and serves as generated non-member data for $\Mtwo$. The new model $\Mtwo$ is thus trained on generated members $G_M$ rather than natural members $N_M$. The new model $\Mtwo$ in turn generates a new dataset $G' = G'_M \cup G'_N$, where $G'_M \cap G'_N = \varnothing$; samples in $G'_M$ may be used to train further downstream models such as $\Mthree$, while $G'_N$ remains withheld. Under this setup, \MGI asks whether a given sample should be attributed to training data or to model-generated data. For the original model $\Mone$, the task is to distinguish among ${N_M, N_N, G}$, separating true natural training members $N_M$ from natural non-members $N_N$ (as in the canonical membership inference task) and from model's $\Mone$ generated samples $G$. For the derivative model $\Mtwo$, the task becomes: distinguish among ${G_M, G_N, G'}$, separating generated training members $G_M$ from both generated non-members $G_N$ and from model's $\Mtwo$ generated samples $G'$. We can further incorporate the natural samples $N$ as $\Mtwo$'s non-member data, however, the $G_N$ represents the most difficult case of the non-member data. 
    }} %
    \label{fig:overview}
    \vspace{-1em}
\end{figure}

Generative models are now trained on massive internet data and generate high-quality samples at an unprecedented speed. These models also inadvertently memorize some of their individual training inputs and later recreate them as outputs~\cite{kowalczuk2025privacyIARs,carlini2023extracting}. The fact that the outputs from generative models are indistinguishable from real data blurs the \textbf{boundary between a model's training and generated data}. 
We formalize this challenge as the Member vs Generated inference (\MGI) task: given an image and a target generative model, decide whether the sample is a true training member of that model or a  generated output by the same model. We illustrate the \MGI task in the overview \Cref{fig:overview} for a \textit{direct training} and a \textit{model derivative} setting. In the \textbf{direct training} setting with model $\Mone$, the goal is to distinguish natural training members $N_M$ from images $G$ generated by the model. Even in this seemingly simple setting, \MGI is fundamentally harder than standard membership inference: generated images are optimized under the same latent distribution as training members, causing their likelihood-based scores to overlap heavily, as we demonstrate in \Cref{sec:limitations_mia}.
We further explore a more challenging and practically relevant \textbf{model derivative} setting, where the samples generated by $\Mone$ are (potentially published online, then scraped from the internet, and) used to train the subsequent model version $\Mtwo$.
In this regime, members are no longer purely natural samples, and simply separating natural from generated content is insufficient. Both membership inference and attribution methods degrade further in the $\Mtwo$ setting, where generated training data introduces compounding ambiguity between membership and generation signals.

Focusing on image generation, we firt show that existing membership inference methods~\cite{kowalczuk2025privacyIARs,yu2025icas,zhai2024clid} are inadequate for MGI: they are designed to separate training members from held-out natural data, and consequently tend to incorrectly label model-generated (but non-member) samples as members. Conversely, attribution methods that aim to determine whether a sample was generated by a particular generative model~\cite{damm2025prada} are also insufficient, often failing by labeling training members as generated. Both failures stem from the same underlying cause: the outputs for the new powerful generative models are derived directly from the training samples of generative models themselves. As a result, signals based on likelihood or output probabilities are similarly high for both true members and the models' own outputs, breaking the assumptions underlying prior methods.

To address the \MGI challenge for modern image generative models, we propose a new method \textit{Data Circuit Breaker} (\DCB).\footnote{A circuit breaker is an electrical safety device designed to protect an electrical circuit from damage caused by current in excess of that which the equipment can handle. In our case, \DCB can protect new models, for example, from degrading in performance by preventing their training on significant amounts of their own generated data.} Our \DCB method treats the generation pipeline holistically rather than focusing solely on the latent generator. The
key insight is that while the latent generator produces high scores for
members and generated samples, the autoencoder introduces measurable artifacts:
generated samples, having passed through the full encode-decode pipeline, exhibit
lower reconstruction and quantization errors than natural data points under the
autoencoder. \DCB exploits this by proceeding in three stages: (1) an
autoencoder-based filtering step that identifies generated samples, separating them from
non-generated data points; (2) a membership inference step on the non-generated samples using the latent generator, where the standard assumption
that members score is restored; and (3) a
cross-generator attribution step that compares conditional log-probabilities
across multiple model versions to distinguish among the generated samples from
different generators. Together, these stages enable \DCB to solve \MGI even in the
most difficult cases of training data memorization.

Overall, our contributions are as follows:
\begin{enumerate}
\item \textbf{New task.} We introduce \emph{Member-vs-Generated Inference} (\MGI) task, which asks whether a given sample is a true training member of a generative model or an output example generated by that same model.
\item \textbf{Limits of prior work.} We demonstrate that existing approaches are insufficient for \MGI: Membership inference methods systematically misclassify generated samples as members, while attribution methods often incorrectly label training members as generated.
\item \textbf{Method.} We propose \DCB (Data Circuit Breaker), a three-stage
procedure that exploits autoencoder self-consistency to filter generated samples,
latent-generator scores for membership inference, and cross-generator probability
discrepancies to trace data circuits across model versions.
\item \textbf{Memorization robustness.} We show that \DCB remains effective even under verbatim memorization, distinguishing original training samples from their regurgitated (near-duplicate) generated counterparts.
\end{enumerate}

\section{Background and Related Work}
\label{sec:background}

\textbf{Image Generative Models (\IGMs).}
The dominant families of modern image generative models (\IGMs) are \emph{diffusion models} (\DMs) and \emph{image autoregressive models} (\IARs). Many state-of-the-art IGMs in both families generate images in a \emph{latent space}: an encoder first maps a high-resolution image from pixel space to a latent representation, and a decoder maps the synthesized latent back to pixels. While they share the latent-generation pipeline, \DMs and \IARs differ fundamentally in how they represent and sample from the data distribution. \DMs define an \emph{implicit} generative process via iterative denoising, whereas \IARs \emph{explicitly} factorize likelihood by predicting token probabilities sequentially, similarly to large language models (LLMs).

\textbf{Diffusion Models (\DMs).} \DMs synthesize images by transforming Gaussian noise into a structured sample through a learned denoising procedure~\cite{song2020,ho2020}. Generation starts from $x_T \sim \mathcal{N}(\mathbf{0}, \mathbf{I})$ and proceeds for $T$ steps, iteratively predicting noise $\epsilon_\gG(\rvx_t, t, \rvc)$ for $t=T,\ldots,1$, and then removing it. In conditional settings (e.g., class-to-image or text-to-image), the denoiser is conditioned on auxiliary inputs $\rvc$, typically text embeddings produced by pretrained encoders such as CLIP~\cite{clip}. Conditioning is injected through cross-attention layers~\cite{vaswani17transformer}.

\textbf{Image Autoregressive Models (\IARs).} \IARs generate images by predicting discrete latent tokens one-by-one using next-token-based autoregressive model, directly modeling a factorized distribution over the latent sequence. A typical \IAR consists of (1) a vector-quantized VAE (VQ-VAE) that encodes an image into discrete representations from a codebook, and (2) an autoregressive transformer that models the codebook representations as tokens and samples them sequentially. For example, LlamaGen~\cite{sun2024autoregressive} uses a VQ-based autoencoder to produce quantized features, then applies a Llama-style transformer to generate tokens autoregressively. VAR further introduces a multi-scale VQ representation to enable coarse-to-fine synthesis~\cite{tian2024var}. Randomized autoregressive models (RARs) generalize next-token prediction by training with randomized token orderings and an annealing-based procedure~\cite{yu2025randomized}.

\textbf{Membership Inference Attack (MIA).} MIA aims to determine whether a given data point was part of a model's training set or not~\cite{shokri2017membership,salem2018ml}. MIA methods are used for auditing models' privacy leakage and verifying empirically the differential privacy guarantees~\cite{marek2026benchmarking,rossi2026natural}. Recent work on MIA against IGMs~\cite{kowalczuk2025privacyIARs,yu2025icas,zhai2024clid} shows that comparing an image's conditional generation to its unconditional generation provides an effective signal for deciding whether the model was trained on that image (member) or not (non-member). Thus, the attack considers only the problem of differentiating between the train vs test samples and does not consider the data generated by the target IGMs. The signal in MIA can be improved by leveraging shadow models, that are trained on data from the same distribution. LiRA~\cite{carlini2022lira} uses the shadow models to estimate the sample's loss distribution for members and non-members, while RMIA~\cite{rmia} compares the likelihood ratio of the target sample with those of reference population samples.

\textbf{Image Attribution Methods.} In contrast to MIAs, image attribution methods seek to identify whether a given image was \textit{generated} by a model or not, which is critical for tracing generated content and preventing data circuits that lead to model collapse~\cite{alemohammad2024selfconsuming,shumailov2024ai}. Analogously to MIAs for image autoregressive models~\cite{kowalczuk2025privacyIARs,yu2025icas}, PRADA~\cite{damm2025prada} shows that the probability ratio can also carry information about whether an image is generated, i.e., a member of the model's learned distribution, or not generated by the target model. However, the  evaluation of the PRADA method is limited to distinguishing generated samples from held-out test samples, which is substantially easier than our newly defined setting: differentiating generated outputs from member training samples. Additionally, PRADA considers only IARs and relies exclusively on the per-token probabilities returned by the image latent generator. As a result, it does not exploit informative signals available in the models' autoencoders, such as the quantization loss between generated and natural (e.g., train or test) samples~\cite{zhao2026data}, leaving part of the membership-related information available in IGMs unexploited.

\textbf{Data Memorization.} Memorization describes the extent to which a model retains information from its training data. It can be \textit{unintended}, when the model stores details about individual examples that can later be reproduced or extracted~\cite{carlini2023extracting,kowalczuk2025privacyIARs}. The \textit{intended} memorization occurs when the model encodes general, reusable patterns that support generalization~\cite{feldman2020does,wang2024memorization}. For data provenance, the most challenging setting arises when a generative model memorizes training images \emph{verbatim} and subsequently regurgitates them during generation, as was shown for \DMs~\cite{carlini2023extracting} and \IARs~\cite{kowalczuk2025privacyIARs}, effectively collapsing the distinction between genuine training images and model-generated outputs. We show that our approach remains effective even in this extreme regime: despite near-duplicate visual content, \IGM samples retain subtle generation-specific residuals that are imperceptible to humans yet detectable in the \IGMs' latent representations, enabling reliable discrimination between natural training images and generated images.

\section{Member vs Generated Inference (\MGI)}

In this section, we formulate \MGI and its threat model under both the direct training and the model derivative settings. 

\textbf{Threat Model.} The threat model of \MGI initially follows that of membership inference and  attribution tasks. Given a set of data points $D$ and access to a generative model $\Mgeneral$, the goal is to differentiate the subset of member samples $D_M$ used for training of the model $\Mgeneral$, its generated samples $D_G$, and non-member samples $D_N$, that were neither used for training, nor generated by the model. We formalize the task for both \textit{direct training} and \textit{model derivative} setting as follows.

\textbf{Direct Training.}
The generative model $\Mone$ was trained on a natural dataset $N_M$, which we refer to as the natural members. Similar to the conventional MIA setting, there is a natural dataset that was not used for training $N_N$, which is the natural non-members. Under \MGI, we also consider the generated data $G$, which was produced by $\Mone$. The goal of \MGI is to distinguish between the generated samples $G$ and members $N_M$. In \Cref{sec:limitations_mia} we show that existing MIAs fail under this new task. \todo{Is this really the only goal? We also distinguish between NN and NM, G and NN. Maybe we say "distinguish between the three possible data sources" or smth.}

\textbf{Model Derivatives.} 
Given the continuous development of generative models and the ubiquity of the generated content, we also consider the relevant scenario of model derivatives. While the samples $N_M$ were used to train the initial model version $\Mone$, its generated samples $G$ may end up being used to train a new model $\Mtwo$, resulting in the set $G_M$, the generated members of $\Mtwo$ and jointly $G_N$ the generated non-members. The second model version $\Mtwo$ produces new samples $G'$ that may end up in further model generations. This iterative training results in data circuits, where new models are derived directly from the previous ones, which can lead to model collapse~\cite{shumailov2024ai,alemohammad2024selfconsuming}. Under the \MGI setting we assume access to both $\Mone$ and $\Mtwo$ and the goal is to distinguish $N_M$ vs $G_M$ vs $G'$. \todo{@Bihe is this correct?}

\textbf{Model Composition.}
A generative model $\gM$ consists of an autoencoder $\gA = \gD\circ\gE$, pairing an encoder $\gE$ with a decoder $\gD$, and a latent generator $\gG$.
$\gM$ can thus be defined as a triplet composition of $\gE, \gD,$ and $\gG$: $ \gM = \langle\gE, \gD, \gG\rangle.$

\section{Limitations of MIA and Attribution Methods}
\label{sec:limitations_mia}

\begin{figure*}[t!]
    \centering
    \begin{subfigure}[b]{0.49\textwidth}
        \centering
        \includegraphics[width=\textwidth]{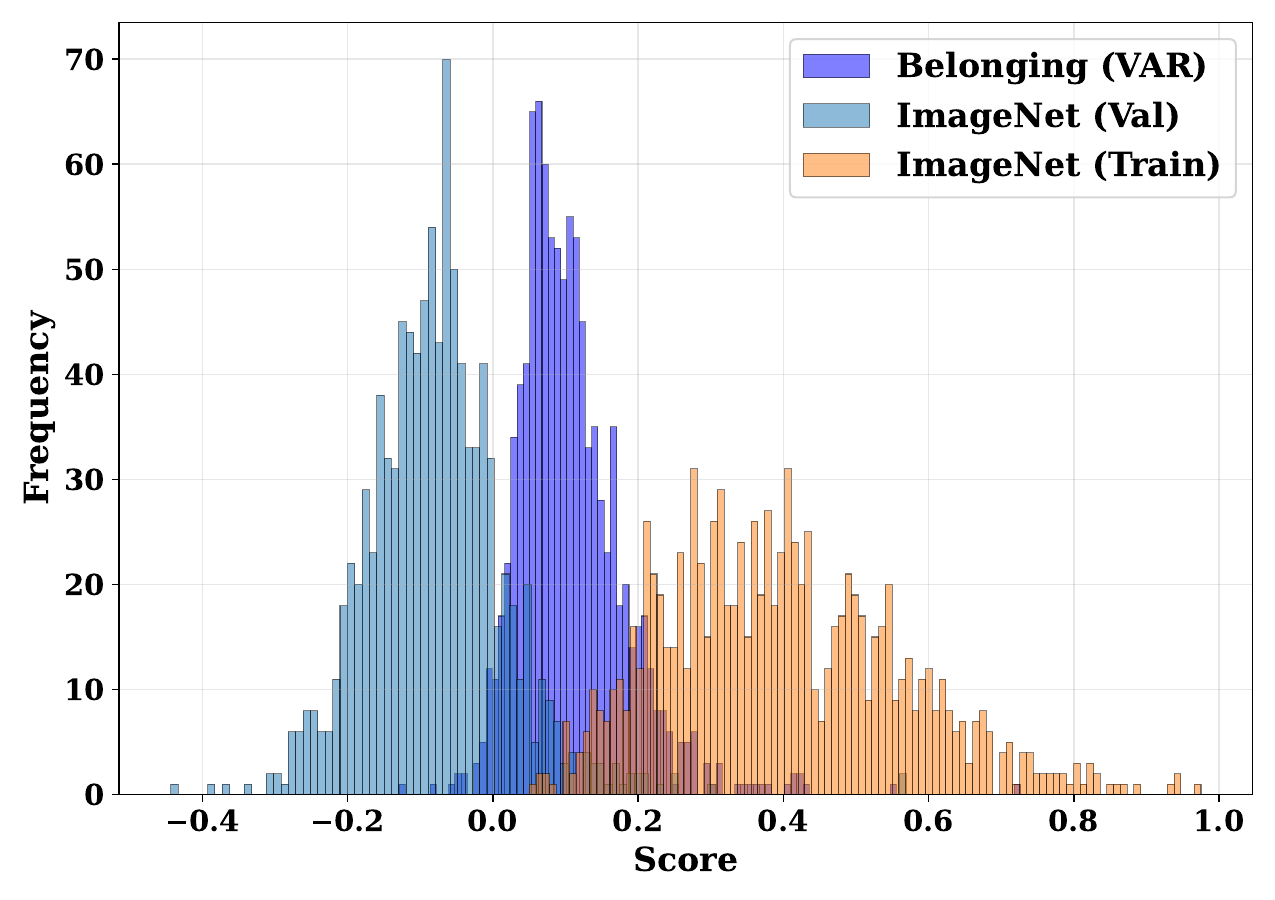}
        \caption{The distribution of scores for the state-of-the-art MIA on IARs~\cite{kowalczuk2025privacyIARs}.}
        \label{fig:var30_delta_fpr}  %
    \end{subfigure}
    \hfill
    \begin{subfigure}[b]{0.49\textwidth}
        \centering
        \includegraphics[width=\textwidth]{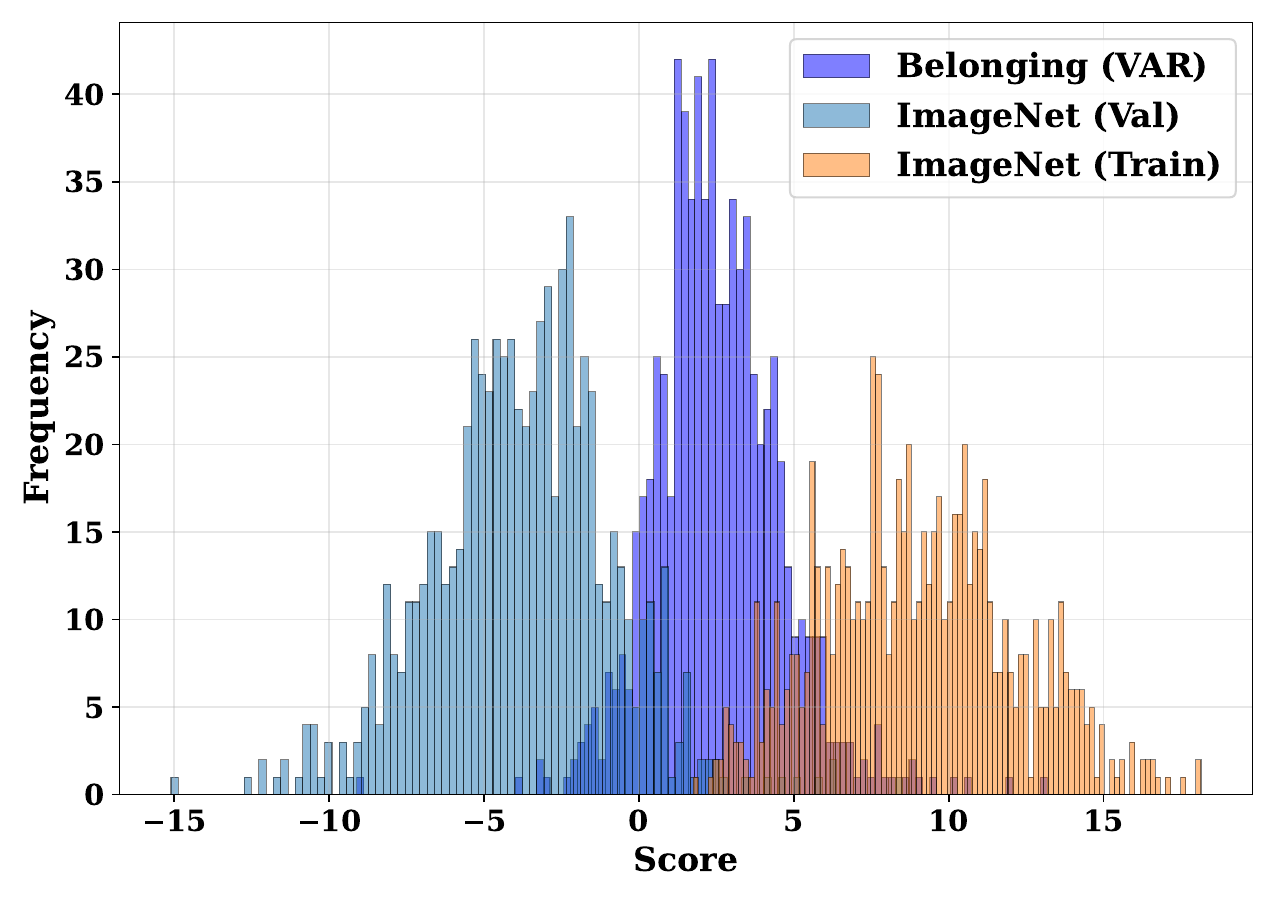}
        \caption{The distribution of scores for a IAR-generated image attribution~\cite{damm2025prada}.}
        \label{fig:var30_prada_fpr}  %
    \end{subfigure}
    \caption{\textbf{Distributions of scores for membership inference attack and image attribution on IARs.} In all cases, the differentiation between training (Train) and generated (Belonging) images is more difficult than between training (Train) and validation (Val) images. This indicates more difficult cases of the MGI (Member vs Generated Inference) than MI task. The evaluated model is VAR~\cite{var_tian2024visualautoregressivemodelingscalable}.
    }
    \label{fig:var30_delta_prada_fpr}
    \vspace{-0em}
\end{figure*}

MIAs for IGMs~\cite{kowalczuk2025privacyIARs,yu2025icas,zhai2024clid} are formulated for the classical setting of distinguishing \emph{natural} training members from \emph{natural} non-members, and they rely almost exclusively on likelihood-based or probability-based signals from the \emph{latent generator}. A representative family of methods score an input image $\rvx$, with conditioning $\rvc$, e.g. a class label or a prompt, via a \emph{conditional probability discrepancy} (CPD): 

\ifcvpr

\begin{align}
\begin{split}
\Delta(\gM,\rvx,\rvc)
& = \log P_{\gM}(\rvx\mid \rvc) - \log P_{\gM}(\rvx)\\
& \approx \log P_{\gG}(\gE(\rvx)\mid \rvc) - \log P_{\gG}(\gE(\rvx)),
\end{split}
\end{align}

\else

\begin{equation}
\Delta(\gM,\rvx,\rvc)
= \log P_{\gM}(\rvx\mid \rvc) - \log P_{\gM}(\rvx)
\approx \log P_{\gG}(\gE(\rvx)\mid \rvc) - \log P_{\gG}(\gE(\rvx)),
\end{equation}

\fi

where the approximation reflects the standard IGM decomposition into an encoder $\gE$ (mapping pixels to latents) and a latent generative model $\gG$ (assigning probabilities in latent space). 
The decision rule is obtained by thresholding $\Delta$, where members are expected to exhibit systematically larger discrepancies than non-members, as the model \textit{remembers} these samples.

This design implicitly treats the autoencoder $\gA$ as a transparent part and largely discards signals that arise from the pixel-to-latent and latent-to-pixel mapping itself. However, the autoencoder $\gA$ is a core component of modern IGMs: the encoder $\gE$ (typically CNN-based) maps an image $\rvx\in\mathbb{R}^{H\times W\times 3}$ to a latent feature map $f\in\mathbb{R}^{\frac{H}{p}\times \frac{W}{p}\times C}$ via down-sampling by a factor $p$, and the corresponding decoder $\gD$ reconstructs $\rvx$ from $f$. As we show later, these components encode artifacts that are not captured in the likelihood-only or probability-only tests on the latent generator $\gG$.

In this paper we consider SOTA MIAs for DMs and IARs, that leverage the CPD. CLiD~\cite{zhai2024clid}, is an MIA for DMs, which approximates the CPD, by leveraging the noise prediction loss to compute the Evidence Lower Bound (ELBO) of the log-likelihood. For IARs we use the MIA from \cite{kowalczuk2025privacyIARs}, which is computed on the model probabilities directly and call this method \textit{PIAR} in the following. Additionally we use ICAS~\cite{yu2025icas}, which considers the classifier-free guidance as an implicit classifier and approximates $p(c|x)$ further weighting this probability to obtain a final score. Furthermore we use PRADA~\cite{damm2025prada}, which while proposed for image attribution, has on a high-level, conceptual similarities to MIAs, as both leverage that models that have \textit{seen} the data, be it during training or generation, have a higher likelihood on that image. PRADA first computes a balanced ratio of the CPD and then uses this ratio in a linear scoring function to obtain a final per-image score. 

\begin{figure*}[t!]
    \centering
    \begin{subfigure}[b]{0.32\textwidth}
        \centering
        \includegraphics[width=\textwidth]{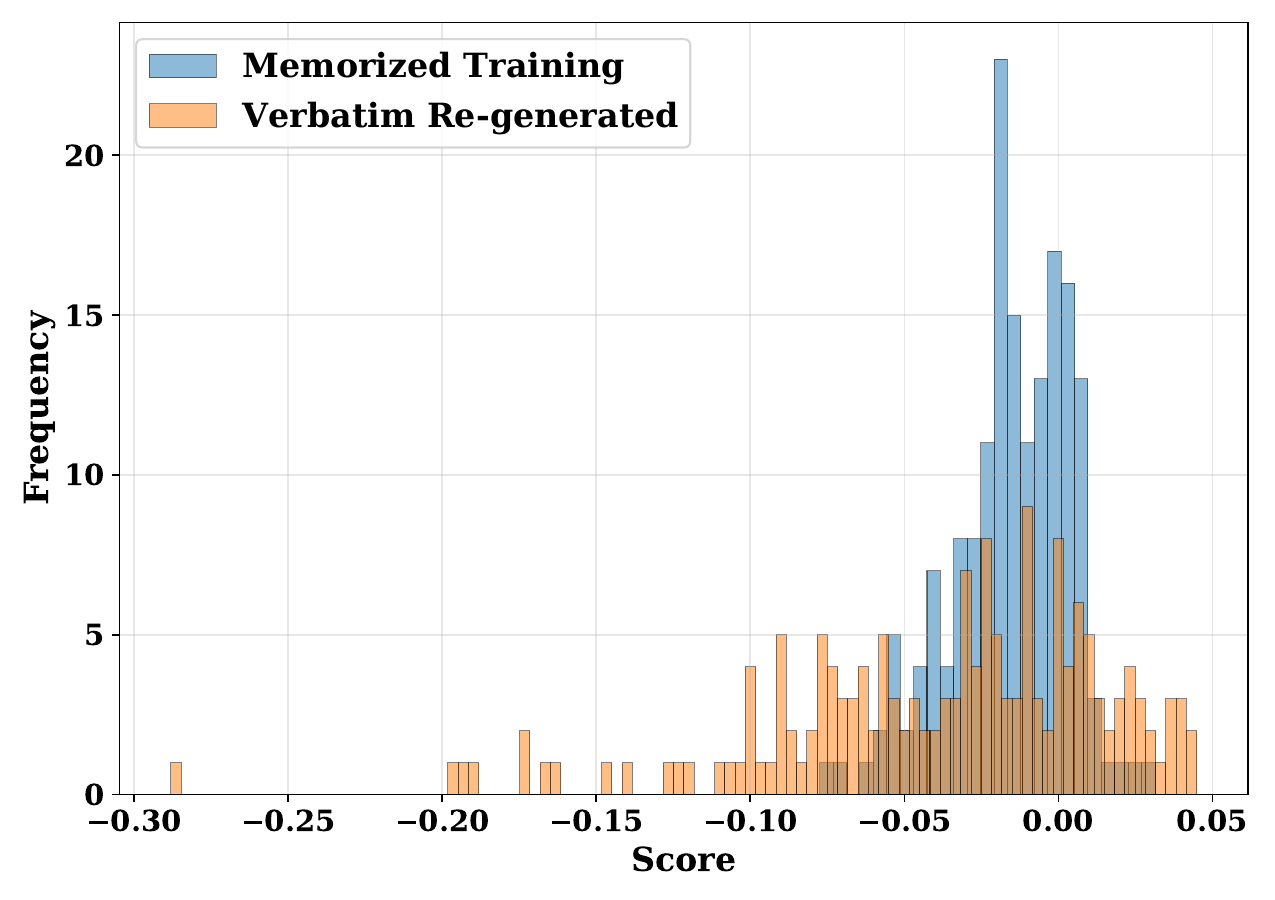}
        \caption{The distribution of scores for the state-of-the-art membership inference on IARs~\cite{yu2025icas}.}
        \label{fig:delta_memorized}  %
    \end{subfigure}
    \hfill
    \begin{subfigure}[b]{0.32\textwidth}
        \centering
        \includegraphics[width=\textwidth]{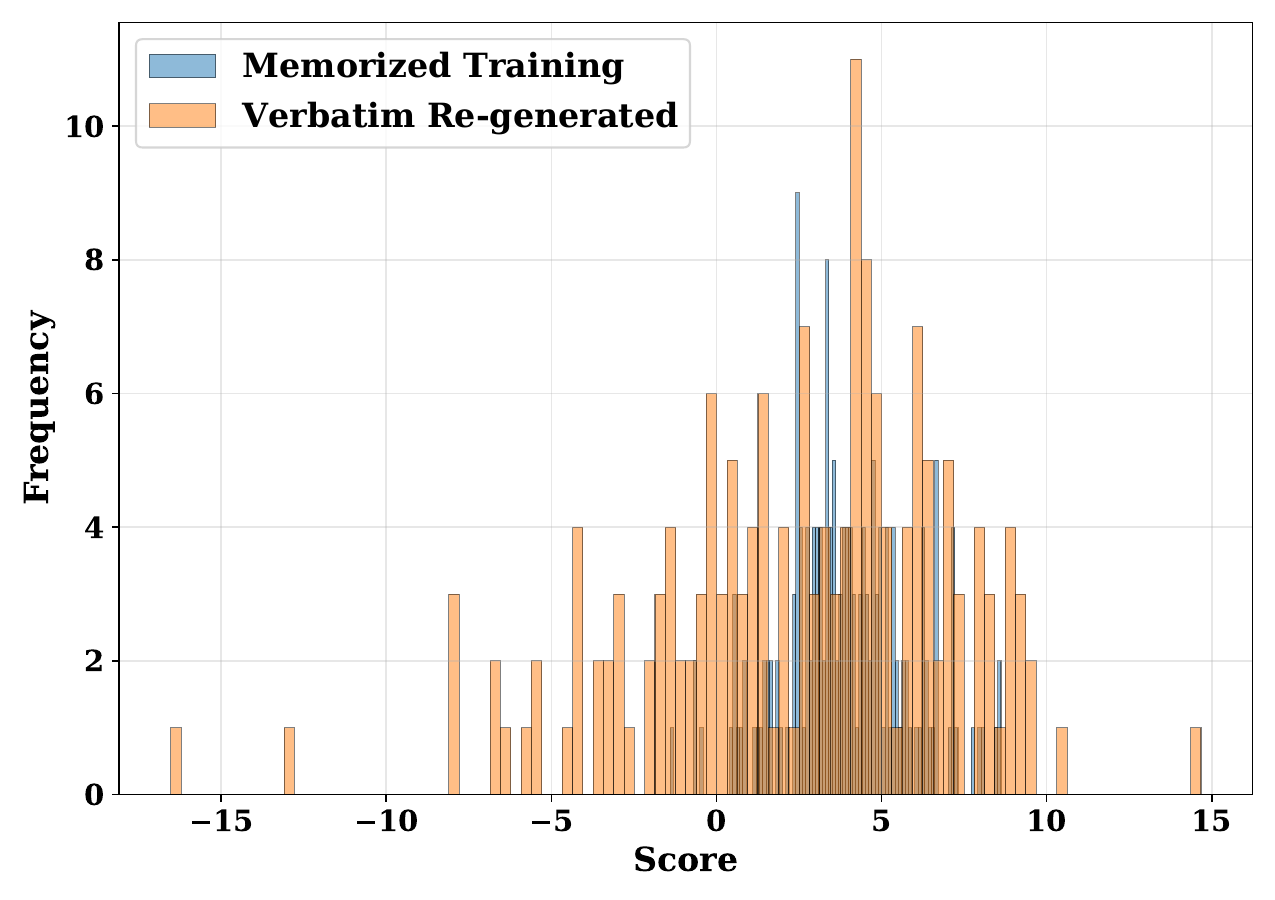}
        \caption{The distribution of scores for a IAR-generated image attribution~\cite{damm2025prada}.}
        \label{fig:prada_memorized}  %
    \end{subfigure}
    \hfill
    \begin{subfigure}[b]{0.32\textwidth}
        \centering
        \includegraphics[width=\textwidth]{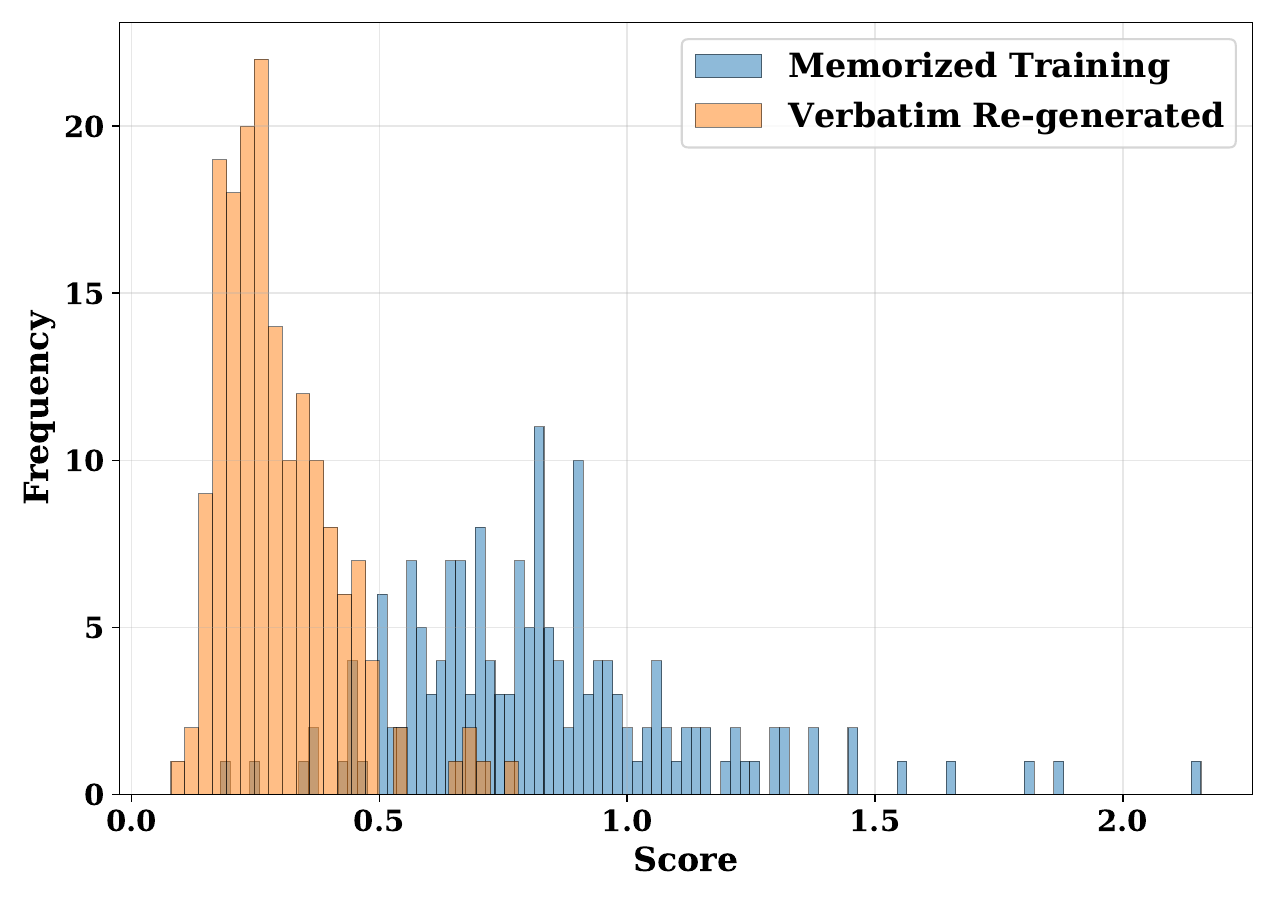}
        \caption{The distribution of scores for the quantization and reconstruction error.}
        \label{fig:ae_memorized}  %
    \end{subfigure}
    \caption{\textbf{Distributions of scores for memorized training samples vs re-generated cases.}  
    State-of-the-art membership inference (a) and attribution (b) methods fail to distinguish memorized training samples (\textit{Memorized Training}) from their verbatim generated counterparts (\textit{Verbatim Re-generated}), whereas our approach (c) clearly separates the two. The evaluated model is RAR-XXL~\cite{yu2025randomized}.
    }
    \label{fig:delta_prada_fpr_memorized}
    \vspace{-0.5em}
\end{figure*}

\subsection{CPD-based Methods Fall Short for \MGI} 
Recent IGM-specific MIAs exploit classifier-free guidance~\cite{ho2022classifier}, where the model is trained (and evaluated) both with conditioning (e.g., class/prompt) and without it, making $\Delta(\cdot)$ a natural statistic to measure if the model \textit{remembers} a specific prompt, image pair. Yet, in the \MGI setting, generated images are \emph{also} optimized to score highly under the same latent generator that produced them. Consequently, $\Delta(\cdot)$ can be simultaneously large for both true members and the model's own outputs, collapsing the separation that MIAs rely on. In short, likelihood-based or probability-based MIAs are well-suited to member vs held-out \emph{natural} data, but they are not designed to distinguish members from \emph{model-generated} non-members, nor do they leverage potentially discriminative signals available in the autoencoder part of IGMs.

\subsection{Case Study: Memorized Training Samples}\label{subsec:mem_case_study}

A special case of our \MGI task arises when the model regenerates images largely resembling the training samples, which is known as \textbf{memorized training samples}. We adopt the methodology proposed by~\cite{kowalczuk2025privacyIARs} to identify 169 memorized training samples for RAR-XXL~\cite{rar_yu2024randomizedautoregressivevisualgeneration}, where each memorized sample features an SSCD~\cite{pizzi2022self} similarity score higher than 0.7.

\ifcvpr
\begin{table}
\else
\begin{wraptable}{r}{0.43\textwidth}
\fi
    \vspace{-10pt} %
  \centering
    \scriptsize
  \caption{\textbf{Performance for different methods for identifying memorized samples.} The evaluated model is RAR-XXL.}
  \label{tab:example}
  \vspace{5pt}
  \resizebox{0.43\textwidth}{!}{
  \begin{tabular}{lcccc}
    \toprule
    Metrics & Delta & ICAS & PRADA & Ours \\
    \midrule
    AUC          & 61.8 & 61.4 & 57.9 & \textbf{97.5} \\
    TPR@5\%FPR   & 3.0  & 0.0  & 0.0  & \textbf{93.5} \\
    \bottomrule
\end{tabular}
  }
  \vspace{-1em}
\ifcvpr
\end{table}
\else
\end{wraptable}
\fi
Concretely, we treat the 169 memorized training images as the \textit{original} samples and their corresponding RAR-XXL outputs as the \textit{generated} samples. The resulting score distributions for membership inference
and image attribution are shown in \Cref{fig:delta_memorized} and \Cref{fig:prada_memorized}, respectively. We observe an even greater overlap between memorized training samples and their regenerated counterparts than in the standard \MGI comparison. Crucially, despite this highly challenging scenario, our approach leverages multiple different attribution signals and reliably separates memorized training images from their generated counterparts and substantially outperforms MIA-based methods, as shown in \Cref{fig:ae_memorized}.
Further results about the memorized training samples can be found in~\Cref{app:mem_samples}.

\section{Proposed Data Circuit Breaker}\label{sec:method}

For our proposed solution to \MGI, we combine signals from (1) the \emph{autoencoder} that maps between pixels and latents and (2) the \emph{latent generator} that models the latent distribution. The key idea is that an image generated by a particular IGM tends to be \emph{more self-consistent} with that model’s autoencoder and latent generator than any natural image or an image generated by a different model, while membership-specific effects (train vs held-out) are more reliably detected after filtering likely-generated samples.

\subsection{Autoencoder Self-Consistency}
\label{sec:ae_signals}

Given the autoencoder $\gA$, where the encoder $\gE$ maps an image $\rvx$ from the pixel-space to a latent representation and the decoder $\gD$ reconstructs the image from the latent representation, we define the reconstruction error:
\begin{equation}
    \recloss(\rvx)
    = \mse\big(\rvx,\gA(\rvx)\big)
    = \mse \big(\rvx, \gD\circ\gE(\rvx)\big),
\end{equation}
where $\mse(\cdot,\cdot)$ is the mean squared error. Following AEDR~\cite{wang2025aedr}, we use a \emph{double reconstruction ratio} to normalize the loss:
\begin{equation}
    \reclossratio(\rvx)
    =
    \frac{\recloss(\rvx)}{\mse\big(\gA(\rvx), \gA\circ\gA(\rvx)\big)}.
\end{equation}
Intuitively, the denominator acts as a per-image baseline: if an image is well-aligned with the autoencoder manifold, the second reconstruction introduces hardly any loss, stabilizing the ratio across diverse content.

\paragraph{VQ-VAE Quantization Error.}
For IARs, the autoencoder is typically a VQ-VAE, which introduces an additional, highly informative signal, namely \emph{quantization error}. The reconstruction procedure of the VQ-VAE introduces a quantization step, where $\gQ$ maps a continuous latent representation of an image $\rvx$ to entries of a codebook. The inverse $\gQinv$, reverts this process and maps codebook indices to the latent representation.
We define the quantization error $\quantloss(\rvx)$ as:
\begin{equation}
\label{eq:quantloss}
    \quantloss(\rvx)
    =
    \mse\big(\gE(\rvx), \gQinv\circ\gQ\circ\gE(\rvx)\big).
\end{equation}

Images synthesized by a given IAR tend to incur smaller $\quantloss$ under that model’s VQ-VAE compared to natural images or images generated by other IGMs, as only they are produced \emph{through} the same discrete codebook. We therefore use the combined autoencoder attribution score
\begin{equation}\label{eq:ae_score}
    \aeloss(\rvx)=
    \begin{cases}
        \reclossratio(\rvx)\cdot \quantloss(\rvx), & \text{(IAR / VQ-VAE)}\\
        \reclossratio(\rvx), & \text{(DM / VAE)}.
    \end{cases}
\end{equation}

\paragraph{Optional Encoder Refinement for IARs.}
The limited alignment between encoder and decoder in IARs introduces additional losses and attribution can degrade. Following~\cite{zhao2026data}, we therefore optionally refine the encoder post hoc by fine-tuning $\hat{\gE}$ to better invert the decoder $\gD$. Fine-tuning is performed on a \emph{disjoint} set of latent feature maps $\rvz$ that were generated by the IAR with the inversion loss:
\begin{equation}
    \invloss
    = \mse\big(\hat{\gE}\circ\gD(\rvz), \rvz\big),
\end{equation}
which improves the stability of $\aeloss$ while preserving the post-hoc setting, as no changes are introduced to the latent generator.

\subsection{Cross-Generator Consistency}
\label{sec:cross_model}
While the autoencoder attribution score is able to identify images that were likely generated by a given model family, they are insufficient to attribute the \emph{exact latent generator}. Especially in the model derivative setting, all generated images are decoded by the same decoder, and the autoencoder attribution score remains identical for all of them. Therefore we introduce an additional generator-based features derived from conditional probability discrepancy.  Given two candidate models $\Mone$ and $\Mtwo$, we form a two-dimensional feature vector based on the conditional probability of the two models.
\begin{equation}
    \rvphi(\rvx,\rvc)=\big(\log P_{\gG_1}(\gE(\rvx)\mid \rvc), \log P_{\gG_2}(\gE(\rvx)\mid \rvc)\big).
\end{equation}
Intuitively this vector encodes the information about the membership of $\rvx$ to both the latent generator $\gG_1$ and $\gG_2$.

As we have access to the image generative models $\Mone$ and $\Mtwo$, for which we want to infer the membership of given samples, we start by estimating the class-conditional densities over $\rvphi$ by generating new data with $\Mone$ and $\Mtwo$. \todo{Bihe could you check this, this paragraph does not make sense to me. Why do we already talk about how the Protocol works? I think we have to remove this and put it into 5.3. I commented the parts out. Please take a look at them. }

We then estimate class-conditional densities over $\rvphi$ using reference samples drawn from each model (e.g., freshly generated sets with independent random seeds) and perform attribution via likelihood comparison (KDE in our implementation). This cross-model view provides separation even when absolute discrepancy values overlap, because images tend to be relatively more \textit{consistent} with the generator that produced them.
We construct reference sets from each source: a reference set $\mathcal{R}_{G}$ drawn from $G$ (representing $\gM_1$-generated images) and a reference set $\mathcal{R}_{G'}$ drawn from $G'$ (representing $\gM_2$-generated images).
We then fit class-conditional KDE densities over the feature vectors of each reference set:

\ifcvpr

\begin{align}\label{eq:kde_stage2}
    \hat{p}_{G}(\rvphi) = \frac{1}{|\mathcal{R}_{G}|}\sum_{i=1}^{|\mathcal{R}_{G}|} K_h\!\big(\rvphi - \rvphi_i^{G}\big),\\
    \hat{p}_{G'}(\rvphi) = \frac{1}{|\mathcal{R}_{G'}|}\sum_{j=1}^{|\mathcal{R}_{G'}|} K_h\!\big(\rvphi - \rvphi_j^{G'}\big).
\end{align}

\else 

\begin{equation}\label{eq:kde_stage2}
    \hat{p}_{G}(\rvphi) = \frac{1}{|\mathcal{R}_{G}|}\sum_{i=1}^{|\mathcal{R}_{G}|} K_h\!\big(\rvphi - \rvphi_i^{G}\big),\qquad
    \hat{p}_{G'}(\rvphi) = \frac{1}{|\mathcal{R}_{G'}|}\sum_{j=1}^{|\mathcal{R}_{G'}|} K_h\!\big(\rvphi - \rvphi_j^{G'}\big).
\end{equation}

\fi

\subsection{Attribution Protocol}
\label{sec:protocol}

We combine the above signals in a cascade designed for the \MGI setting.

\paragraph{Stage 1: Autoencoder-based Filtering (Generated vs. Non-Generated).}
We first apply the autoencoder score $\aeloss(\rvx)$ to identify a high-confidence subset of \emph{IGM-generated} samples and separate them from samples that are unlikely to be produced by the target pipeline. 

\paragraph{Stage 2: Membership Inference on the Remaining Samples (Member vs. Non-Member).}
On images detected non-generated by Stage 1, we apply standard MIA-style scoring based on the latent generator (e.g., $\Delta(\gM,\rvx,\rvc)$ or ICAS) to distinguish training members from non-members. Restricting MIAs to this subset restores their core assumption (members vs non-members), and substantially reduces false positives caused by model-generated images. With stage 1 and 2, we address the \MGI problem for the direct training setting of $\Mone$.
Specifically, we instantiate our second stage with ICAS, which is the best-performing MIA for most models. We note that, although PRADA outperforms ICAS on RAR, it requires an extra calibration set. This is an extra advantage beyond our main setting, and restricts the applicability of PRADA. Therefore, we do not choose ICAS instead of PRADA to instantiate our stage 2 for any models.

\paragraph{Stage 3: Source Attribution among Generators (Data Circuits).}
In the $M_2$ setting of \Cref{fig:overview}, where training data may itself be generated, we additionally apply cross-model generator attribution using $\rvphi(\rvx,\rvc)$ and reference sets from $\Mone$ and $\Mtwo$ to separate $\Mone$-generated samples used for training $\Mtwo$, $\Mone$-generated samples not used for training $\Mtwo$, and $\Mtwo$-generated samples. Combined with Stage 1 and Stage 2, this yields a practical decomposition into (1) natural members, (2) natural non-members, (3) generated samples attributed to a specific model, and (4) generated samples used for training downstream models, thus fully addressing \MGI in the presence of data circuits.

\section{Empirical Evaluation}

\subsection{Experimental Setup}

\textbf{Models.} We evaluate SOTA IARs and DMs, following the previous work on MIAs for IGMs~\cite{kowalczuk2025privacyIARs,dubinski2024cdicopyrighteddataidentification,zhai2024clid,yu2025icas}. 
Our selection of models requires access to their training sets for our analysis to verify the outcome of MIAs and MGIs.
We use \textbf{\textit{VAR-d30}} (\textit{d} = model depth)~\cite{var_tian2024visualautoregressivemodelingscalable}, \textbf{\textit{RAR-XXL}}~\cite{rar_yu2024randomizedautoregressivevisualgeneration}, and \textbf{\textit{LlamaGen-XXL}}~\cite{sun2024autoregressive}, trained for class-conditioned generation. We download the pre-trained weights from the corresponding repositories and for generation we follow the settings recommended in the original works. For the DMs we focus on the UNet~\cite{unet} based architectures \textbf{\textit{Stable Diffusion 1.4}} and \textbf{\textit{2.1}}~\cite{rombach2022high}.

\textbf{Datasets.} As the above IARs were trained on ImageNet-1k~\cite{deng2009imagenet} dataset, we use it to perform our MGI and MIA tasks. We sample 1000 samples from the training set as members and similarly 1000 samples from the validation set as non-members. For Stable Diffusion we follow CLiD~\cite{zhai2024clid} and first fine-tune the model on a set of 2500 MS-COCO images for 50k steps to obtain a set of natural member samples and non-member samples. Then we use 1000 samples from training as members and 1000 samples from validation as non-members.

\textbf{Fine-tuning.}
We fine-tune the second model $\Mtwo$ on 5000 images generated by the first model $\Mone$. We denote the images generated by $\Mone$ as $G_M$. We also keep another 1000 images generated by $\Mone$ as a held-out set (denoted as $G_N$, which are \textit{generated} data points that act as \textit{non-members}). For IARs we fine-tune $\Mtwo$ for 5 epochs, while for DMs we use 20. The learning rate is $1\times10^{-5}$ for all models. We provide the hyperparameter details in \Cref{app:implementation_details}.

\textbf{Baselines.}
\bihe{MGI is a \emph{newly defined task} without an existing solution. We follow standard practice for newly defined tasks by adapting SoTA methods from the closest domains MIA and image attribution.}
\bihe{Regarding the \textbf{\textit{MIA baselines}}, we choose SoTA MIA methods for IARs and DMs, respectively. }
For IARs, we use \cite{kowalczuk2025privacyIARs} and refer to the method as PIAR, and ICAS~\cite{yu2025icas}. For the DMs, we use the SOTA MIA CLiD and extend the IAR-based ICAS to DMs based on the CLiD scores. 
In \Cref{sec:strong_mia}, We further test strong MIAs, LiRA/RMIA, which we give an \emph{advantage} by training shadow models.
For the direct training setting, MIAs make the assumption that members have a higher score than all non-member samples and we extend this assumption to \MGI.
\bihe{Regarding the \textbf{\textit{image attribution baseline}}, }we consider PRADA~\cite{damm2025prada}, which is originally proposed for IARs but extended to DMs by us. Under the direct training setting, the image attribution methods make the assumption that generated samples will have the highest score, followed by members and non-members. The same intuition extends to the derivative setting.

\textbf{Metrics.} In the following we focus on the, especially for inference tasks, relevant metric of \tpr and additionally provide the AUC in \Cref{app:auc_results}.

\subsection{Evaluation on the Direct Training Setting}

\ifcvpr
\begin{table*}[t!]
\else
\begin{table}[t!]
\fi
    \centering
    \caption{\textbf{\tpr for IARs in the direct training setting.} Only \DCB achieves consistent performance across all comparisons and models.}
    \label{table:m1_tpr}
    \ifcvpr
    \resizebox{\textwidth}{!}{
    \else
    \tiny
    \fi
    \begin{tabular}{lc*{11}{c}c}
        \toprule
        \multirow{2}{*}{Method} & \multicolumn{4}{c}{RAR} & \multicolumn{4}{c}{VAR} & \multicolumn{4}{c}{LlamaGen} & \multirow{2}{*}{Overall} \\
        \cmidrule(lr){2-5} \cmidrule(lr){6-9} \cmidrule(lr){10-13}
        & $N_M$/$G$ & $N_N$/$G$ & $N_M$/$N_N$ & Avg & $N_M$/$G$ & $N_N$/$G$ & $N_M$/$N_N$ & Avg & $N_M$/$G$ & $N_N$/$G$ & $N_M$/$N_N$ & Avg & \\
        \midrule
        PIAR & 0.0 & 99.5 & 62.6 & 54.0 & 58.6 & 11.5 & 91.7 & 53.9 & 0.5 & 17.7 & 6.7 & 8.3 & 38.8 \\
        ICAS & 0.0 & 99.7 & 72.5 & 57.4 & 61.9 & 33.9 & \textbf{98.7} & 64.8 & 0.0 & 89.6 & \textbf{17.2} & 35.6 & 52.6 \\
        PRADA & 62.7 & \textbf{100.0} & \textbf{81.3} & 81.3 & 0.0 & 24.8 & 96.9 & 40.6 & 9.3 & 68.8 & 7.1 & 28.4 & 50.1 \\
        Ours & \textbf{99.9} & 99.9 & 72.5 & \textbf{90.8} & \textbf{99.3} & \textbf{99.5} & \textbf{98.7} & \textbf{99.2} & \textbf{100.0} & \textbf{100.0} & \textbf{17.2} & \textbf{72.4} & \textbf{87.4} \\
        \bottomrule
    \end{tabular}
    \ifcvpr
    }
    \fi
\ifcvpr
\end{table*}
\else
\end{table}
\fi

\ifcvpr
\begin{table*}[t!]
\else
\begin{table}[t!]
\fi
    \centering
    \caption{\textbf{\tpr for DMs in the direct training setting.} Only \DCB achieves consistent performance across all comparisons and models.}
    \label{table:dm_m1_tpr}
    \ifcvpr
    \resizebox{0.75\textwidth}{!}{
    \else
    \tiny
    \fi
    \begin{tabular}{lc*{11}{c}c}
        \toprule
        \multirow{2}{*}{Method} & \multicolumn{4}{c}{SD1.4} & \multicolumn{4}{c}{SD2.1} & \multirow{2}{*}{Overall} \\
        \cmidrule(lr){2-5} \cmidrule(lr){6-9}
        & $N_M$/$G$ & $N_N$/$G$ & $N_M$/$N_N$ & Avg & $N_M$/$G$ & $N_N$/$G$ & $N_M$/$N_N$ & Avg & \\
        \midrule
        CLiD & 0.0 & 88.2 & \textbf{36.2} & 41.5 & 0.0 & 82.2 & \textbf{31.5} & 37.9 & 39.7 \\
        ICAS & 0.0 & 87.8 & 35.7 & 41.2 & 0.0 & 82.1 & \textbf{31.5} & 37.9 & 39.5 \\
        PRADA & 0.7 & 0.4 & 0.4 & 0.5 & 0.7 & 0.3 & 0.4 & 0.4 & 0.5 \\
        Ours & \textbf{99.9} & \textbf{99.8} & 35.7 & \textbf{78.5} & \textbf{100.0} & \textbf{100.0} & \textbf{31.5} & \textbf{77.2} & \textbf{77.8} \\
        \bottomrule
    \end{tabular}
    \ifcvpr
    }
    \fi
\ifcvpr
\end{table*}
\else
\end{table}
\fi

First we focus on the direct training setting, known from the MIA task, where the model $\Mone$ was trained on natural images resulting in the natural members $N_M$ and natural non-members $N_N$. However, our \MGI introduces the models generated samples $G$ as a new an important part of this task.

Under this new setting, we analyze the performance of the original MIAs and image attribution methods for IARs in \Cref{table:m1_tpr} and report the \tpr. We find that while existing MIAs are able to separate the natural members from the natural non-members, they break when the generated data is introduced. Our \DCB however achieves near 100\% \tpr under the generated vs natural setting and improves the average performance by over 36\% (LlamaGen). Under the \MGI task \DCB benefits from combining multiple signals leading to a consistent performance across detections.  

We additionally analyze the \MGI task on DMs in \Cref{table:dm_m1_tpr}, following CLiD~\cite{zhai2024clid} and fine-tuning the models on natural MS-COCO data to obtain natural members and non-members. Our results show that, similar as for the IARs, while the baseline methods are able to distinguish $N_N$ and $N_M$, they fail when the generated data is introduced. This difficulty for MIAs is additionally visualized in \Cref{app:visualization}, which plots the score distributions for the different datasets. As the generated data was produced by $\Mone$, the MIA obtains a high score, as the model \textit{remembers} the sample. This occurs because likelihood-based scores are similarly elevated for both member samples and the model's own outputs, collapsing the separation that MIAs rely upon. Only \DCB, which takes the full generative pipeline into account, can distinguish the generated data from the natural data. This effect is especially pronounced in the $\Mtwo$ setting, where \DCB beats the baselines by more than 39\% (SD2.1).

\subsection{Evaluation on the Model Derivative Setting}

As images generated by IGMs experience a widespread reuse, we shift the focus to the derivative setting, where a model $\Mtwo$ was fine-tuned on generated images $G_M$, which are now the member samples of $\Mtwo$. The new model continuously generates new images $G'$, which, with the generated non-members $G_N$ introduces three datasets to the \MGI task. In \Cref{table:m2_tpr} we report the \tpr for distinguishing the different datasets and find that the existing methods struggle to differentiate the distributions for both IARs and DMs. Our \DCB, on the other hand, is able to clearly distinguish the two sets.

The difficulty of this new \MGI setting is reflected in the comparison of \Cref{table:m2_tpr}. While the baselines achieve reasonable performance for most \textit{Natural vs Generated} cases, they struggle when differentiating within the generated samples. Particularly for the DMs, the detection performance collapses. The score distributions in \Cref{fig:m2_mia} and \Cref{app:visualization} provide additional insights as to why the \MGI setting is fundamentally more difficult. Contrary to MIA assumption, the score of the generated samples is larger than the score of non-members and overlaps or exceeds the score of the members. This property of the generated samples is why standard MIA fail.

\ifcvpr
\begin{table*}[t!]
\else
\begin{table}[t!]
\fi
    \centering
    \caption{\textbf{\tpr for the model derivative setting.} Most existing methods fail to attribute generated samples correctly.}
    \label{table:m2_tpr}
    \tiny
    \resizebox{\textwidth}{!}{
    \begin{tabular}{llc*{9}{c}c}
        \toprule
        \multirow{2}{*}{Model} & \multirow{2}{*}{Method} & \multicolumn{6}{c}{Natural vs Generated} & \multicolumn{3}{c}{Among Generated}  & \multicolumn{1}{c}{Natural} & \multirow{2}{*}{Overall} \\
        \cmidrule(lr){3-8}\cmidrule(lr){9-11}\cmidrule(lr){12-12}
        & & $N_M$/$G_M$ & $N_M$/$G_N$ & $N_M$/$G'$ & $N_N$/$G_M$ & $N_N$/$G_N$ & $N_N$/$G'$
        & $G_M$/$G_N$ & $G_M$/$G'$ & $G_N$/$G'$
        & $N_M$/$N_N$
        & \\
        \midrule
        \multirow{4}{*}{VAR}
        & PIAR & 70.6 & 0.2 & 1.9 & 99.9 & 62.4 & 97.8 & 94.0 & 62.8 & 15.8 & 79.2 & 58.5 \\
        & ICAS & \textbf{99.8} & 10.3 & 82.8 & \textbf{100.0} & 89.9 & \textbf{99.8} & 99.6 & 91.8 & 51.6 & \textbf{87.0} & 81.3 \\
        & PRADA & 93.2 & 0.3 & 9.7 & 99.9 & 68.0 & 98.7 & 95.7 & 0.0 & 16.0 & 82.3 & 56.4 \\
        \cdashline{2-13}
        \addlinespace[2pt]
        & Ours & 99.3 & \textbf{99.3} & \textbf{99.4} & 99.5 & \textbf{99.5} & 99.6 & \textbf{100.0} & \textbf{99.0} & \textbf{75.0} & \textbf{87.0} & \textbf{95.8} \\
        \midrule
        \multirow{4}{*}{RAR}
        & PIAR & \textbf{100.0} & 59.9 & 97.0 & \textbf{100.0} & 99.8 & \textbf{100.0} & 74.6 & 18.5 & 16.4 & 62.6 & 72.9 \\
        & ICAS & \textbf{100.0} & 82.9 & 99.2 & \textbf{100.0} & 99.6 & \textbf{100.0} & 95.6 & 20.9 & 49.9 & 72.5 & 82.1 \\
        & PRADA & \textbf{100.0} & 82.1 & 98.0 & \textbf{100.0} & \textbf{100.0} & \textbf{100.0} & 82.4 & 0.0 & 18.9 & \textbf{82.0} & 76.3 \\
        \cdashline{2-13}
        \addlinespace[2pt]
        & Ours & 99.9 & \textbf{99.9} & \textbf{99.9} & 99.9 & 99.9 & 99.9 & \textbf{100.0} & \textbf{100.0} & \textbf{95.6} & 72.5 & \textbf{96.7} \\
        \midrule
        \multirow{4}{*}{SD1.4}
        & CLiD & 84.8 & 71.7 & 98.5 & \textbf{99.9} & 99.6 & \textbf{100.0} & 5.4 & 1.0 & 37.9 & \textbf{36.2} & 63.5 \\
        & ICAS & 84.8 & 71.6 & 98.6 & \textbf{99.9} & 99.6 & \textbf{100.0} & 5.4 & 1.0 & 38.1 & 35.7 & 63.5 \\
        & PRADA & 0.1 & 0.5 & 2.4 & 0.1 & 0.0 & 2.3 & 0.7 & 4.1 & 3.5 & 0.0 & 1.4 \\
        \cdashline{2-13}
        \addlinespace[2pt]
        & Ours & \textbf{99.9} & \textbf{99.9} & \textbf{98.7} & 99.8 & \textbf{99.8} & 98.5 & \textbf{56.0} & \textbf{42.2} & \textbf{95.8} & 35.7 & \textbf{82.6} \\
        \midrule
        \multirow{4}{*}{SD2.1}
        & CLiD & 86.8 & 74.2 & 99.5 & 99.9 & 99.5 & \textbf{100.0} & 5.8 & 0.0 & 47.2 & \textbf{31.5} & 64.4 \\
        & ICAS & 86.8 & 73.9 & 99.5 & 99.9 & 99.5 & \textbf{100.0} & 5.8 & 0.0 & 47.0 & \textbf{31.5} & 64.4 \\
        & PRADA & 0.3 & 0.4 & 0.4 & 0.0 & 0.3 & 0.1 & 0.9 & 2.1 & 1.7 & 0.1 & 0.6 \\
        \cdashline{2-13}
        \addlinespace[2pt]
        & Ours & \textbf{100.0} & \textbf{100.0} & \textbf{99.7} & \textbf{100.0} & \textbf{100.0} & 99.8 & \textbf{52.8} & \textbf{54.0} & \textbf{99.0} & \textbf{31.5} & \textbf{83.7} \\
        \bottomrule
    \end{tabular}

    }
\ifcvpr
\end{table*}
\else
\end{table}
\fi

\subsection{Analysis for Strong MIA}\label{sec:strong_mia}

\ifcvpr
\begin{table*}[t!]
\else
\begin{table}[t!]
\fi
    \centering
    \caption{\textbf{\tpr for the strong MIA.} We consider both LiRA and RMIA and train 5 shadow models.}
    \label{table:m2_strong_mia}
    \tiny
    \resizebox{\textwidth}{!}{
    \begin{tabular}{llc*{9}{c}c}
        \toprule
        \multirow{2}{*}{Model} & \multirow{2}{*}{Method} & \multicolumn{6}{c}{Natural v.s. Generated} & \multicolumn{3}{c}{Among Generated}  & \multicolumn{1}{c}{Natural} & \multirow{2}{*}{Overall} \\
        \cmidrule(lr){3-8}\cmidrule(lr){9-11}\cmidrule(lr){12-12}
        & & $N_M$/$G_M$ & $N_M$/$G_N$ & $N_M$/$G'$ & $N_N$/$G_M$ & $N_N$/$G_N$ & $N_N$/$G'$
        & $G_M$/$G_N$ & $G_M$/$G'$ & $G_N$/$G'$
        & $N_M$/$N_N$
        & \\
        \midrule
        \multirow{3}{*}{VAR}
        & LiRA & 98.7 & 1.0 & 16.7 & 98.7 & 1.1 & 16.8 & 98.7 & 50.3 & 16.7 & 1.1 & 40.0 \\
        & RMIA & \textbf{100.0} & 3.6 & 78.3 & \textbf{100.0} & 77.5 & \textbf{99.6} & 99.9 & 98.3 & 67.2 & 40.0 & 76.4 \\
        \cdashline{2-13}
        \addlinespace[2pt]
        & Ours & 99.3 & \textbf{99.3} & \textbf{99.4} & 99.5 & \textbf{99.5} & \textbf{99.6} & \textbf{100.0} & \textbf{99.0} & \textbf{75.0} & \textbf{87.0} & \textbf{95.8} \\
        \midrule
        \multirow{3}{*}{RAR}
        & LiRA & 92.8 & 0.8 & 31.2 & 95.0 & 1.2 & 36.9 & 94.2 & 18.4 & 34.4 & 1.3 & 40.6 \\
        & RMIA & \textbf{99.9} & 65.7 & 99.5 & \textbf{100.0} & 96.2 & \textbf{100.0} & 99.0 & 77.5 & 71.3 & 17.8 & 82.7 \\
        \cdashline{2-13}
        \addlinespace[2pt]
        & Ours & \textbf{99.9} & \textbf{99.9} & \textbf{99.9} & 99.9 & \textbf{99.9} & 99.9 & \textbf{100.0} & \textbf{100.0} & \textbf{95.6} & \textbf{72.5} & \textbf{96.7} \\
        \bottomrule
    \end{tabular}
    }
\ifcvpr
\end{table*}
\else
\end{table}
\fi

We expand on the explored MIAs, by analyzing stronger methods and employ both LiRA~\cite{carlini2022lira} and RMIA~\cite{rmia} on the model derivative setting and show that even under access to trained shadow models, the MIA fail in the \MGI setting. Both LiRA and RMIA require a scaler score to compute a one-dimensional probability distribution. We use the strongest probability-based MIA method ICAS~\cite{yu2025icas} to convert the token-wise probabilities predicted by IARs into a score scalar for each sample.

Concretely, we obtain 5 shadow models, by fine-tuning $\Mone$ for 5 epochs, with the same hyperparameters used for $\Mtwo$, on datasets of 2500 samples randomly drawn from a 5000-sample shadow dataset. For RMIA, we utilize an additional population dataset of 1000 samples generated by $\Mone$ and set its core hyperparameters to $\alpha=0.3$. This setup enables the methods to estimate the distribution of generated members and non-member samples, giving these methods a strict advantage for the $G_N$ vs $G_M$ case compared to \DCB. We report the \tpr in \Cref{table:m2_strong_mia}, for VAR and RAR. The results highlight that even under significant advantages, strong MIAs are not sufficient to solve the \MGI task. Specifically for the \textit{Natural vs. Generated} identification the strong MIA perform similar to the MIAs without shadow models. Notably, our \DCB consistently outperforms both LiRA and RMIA across all comparisons, highlighting that utilizing the full model pipeline provides stronger signals.

\section{Conclusions}
\label{sec:conclusions}

We introduced Member vs Generated Inference (\MGI), a new and strictly harder inference task than standard membership inference. \MGI requires separating a generative model's training members from its own generated outputs, including in data-circuit settings where subsequent models are trained on generated data. We showed that existing membership inference and attribution methods are inadequate for \MGI because modern generative models produce non-member samples that are closely tied to the training distribution, leading to MIAs systematically misclassifying generated samples as members, while attribution methods mislabel true training members as generated. To address this, we proposed \DCB, a multi-stage pipeline that covers the full generation process by leveraging complementary signals from the autoencoder and latent generator. By first identifying synthesized content and then distinguishing remaining training members from non-members, \DCB is able to consistently outperform previous methods. We demonstrated that \DCB remains effective even on memorization, separating original training samples from their regurgitated counterparts and enabling practical mitigation of harmful data circuits. Finally, we showed that \DCB achieves better detection rate than strong membership inference attacks such as LiRA and RMIA, highlighting that a holistic procedure of the full generative pipeline is essential to solve \MGI.

\section*{Acknowledgments}
This research was funded by the Deutsche Forschungsgemeinschaft (DFG, German Research Foundation), Project number 550224287. Franziska Boenisch received funding from the European Research Council (ERC) under the European Union’s Horizon Europe research and innovation programme (grant agreement No 101220235). We would like to acknowledge our sponsors, who support our research with financial and in-kind contributions: OpenAI and G-Research. We also thank members of the SprintML group for their
feedback. Responsibility for the content of this publication lies with the authors.

{
    \small
    \bibliographystyle{ieeenat_fullname}
    \bibliography{main}
}

\appendix
\crefalias{section}{appendix}
\renewcommand{\tablename}{Table}
\setcounter{table}{0}
\renewcommand{\thetable}{A\arabic{table}}
\renewcommand{\figurename}{Figure}
\setcounter{figure}{0}
\renewcommand{\thefigure}{A\arabic{figure}}

\section{Further Implementation Details}\label{app:implementation_details}

\subsection{Data Pre-processing}
We follow the augmentations in the original training recipes of each model to ensure that our evaluation faithfully reflects the conditions under which membership and generation signals arise.
For VAR and LlamaGen, when computing the MIA scores on $\Mone$, we apply the same data augmentations used during training: each image is first resized by a factor of $1.125$ for VAR and $1.1$ for LlamaGen, followed by a center crop to the model's native resolution ($256\times256$ for VAR and $384\times 384$ for LlamaGen).

\subsection{Fine-tuning}

We provide the hyperparameters for fine-tuning $\Mtwo$ in the model derivative setting in \Cref{table:finetuning_params}.
For all models, we fine-tune exclusively the \emph{latent generator} while keeping the autoencoder weights frozen.
The latent generator corresponds to the transformer in IARs and the UNet in the diffusion models.
This design choice mirrors the common practice in which downstream practitioners adapt only the generative backbone to new data.
The fine-tuning data consists of $5{,}000$ images generated by $\Mone$.
For the IARs (VAR and RAR), $5$ epochs suffice to adapt the latent generator to the generated distribution, whereas for the diffusion models (SD~1.4 and SD~2.1) we train for $20$ epochs due to their slower convergence.
All experiments use a fixed learning rate of $1\times10^{-5}$ with the AdamW optimizer.

\begin{table}[h]
    \centering
    \caption{Hyperparameters for fine-tuning $\Mtwo$ for the derivative setting.}
    \label{table:finetuning_params}
    \footnotesize
    \begin{tabular}{lcccccccc}
        \toprule
        Model & Batch Size & Learning Rate & Training Samples & Epochs \\
        \midrule
        VAR & 4 & $1\times 10^{-5}$ & 5000 & 5 \\
        RAR & 4 & $1\times 10^{-5}$& 5000 & 5 \\
        SD 1.4 & 4 & $1\times 10^{-5}$ & 5000 & 20 \\
        SD 2.1 & 4 & $1\times 10^{-5}$ & 5000 & 20 \\
        \bottomrule
    \end{tabular}
\end{table}

\section{Distribution Visualization on More Models}\label{app:visualization}

In this section, we complement the distribution analysis of the main paper with visualizations for additional models and settings.
These plots substantiate our central observation: across all evaluated architectures, the distributions for generated images overlap heavily with those for training members, making the MGI task fundamentally harder than standard membership inference.

\subsection{Model Direct Training Setting}

\begin{figure*}[t]
    \centering
    \begin{subfigure}[b]{0.49\textwidth}
        \centering
        \includegraphics[width=\textwidth]{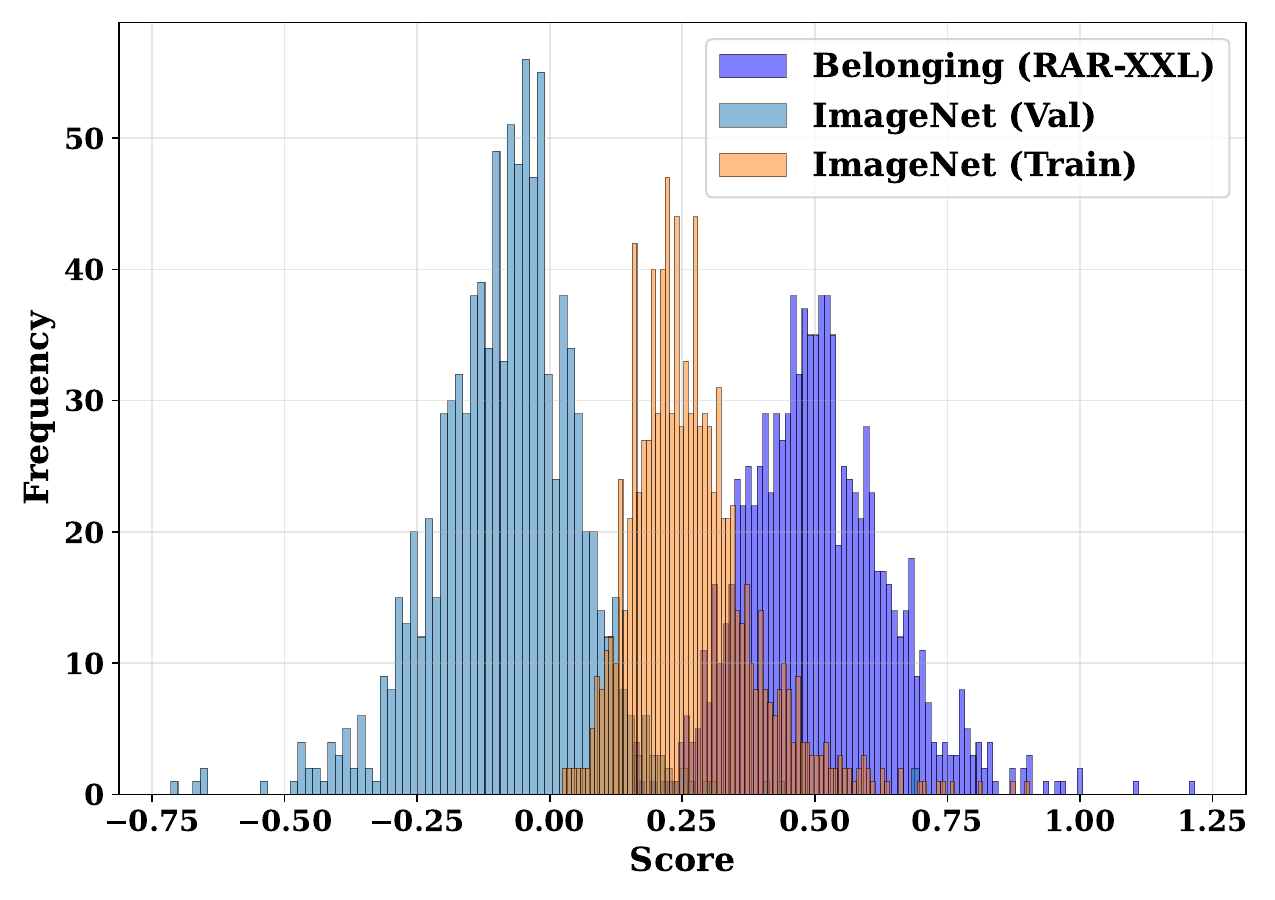}
        \caption{The distribution of scores for the membership inference attack PIAR~\cite{kowalczuk2025privacyIARs}}
        \label{fig:delta_fpr_rar}  %
    \end{subfigure}
    \hfill
    \begin{subfigure}[b]{0.49\textwidth}
        \centering
        \includegraphics[width=\textwidth]{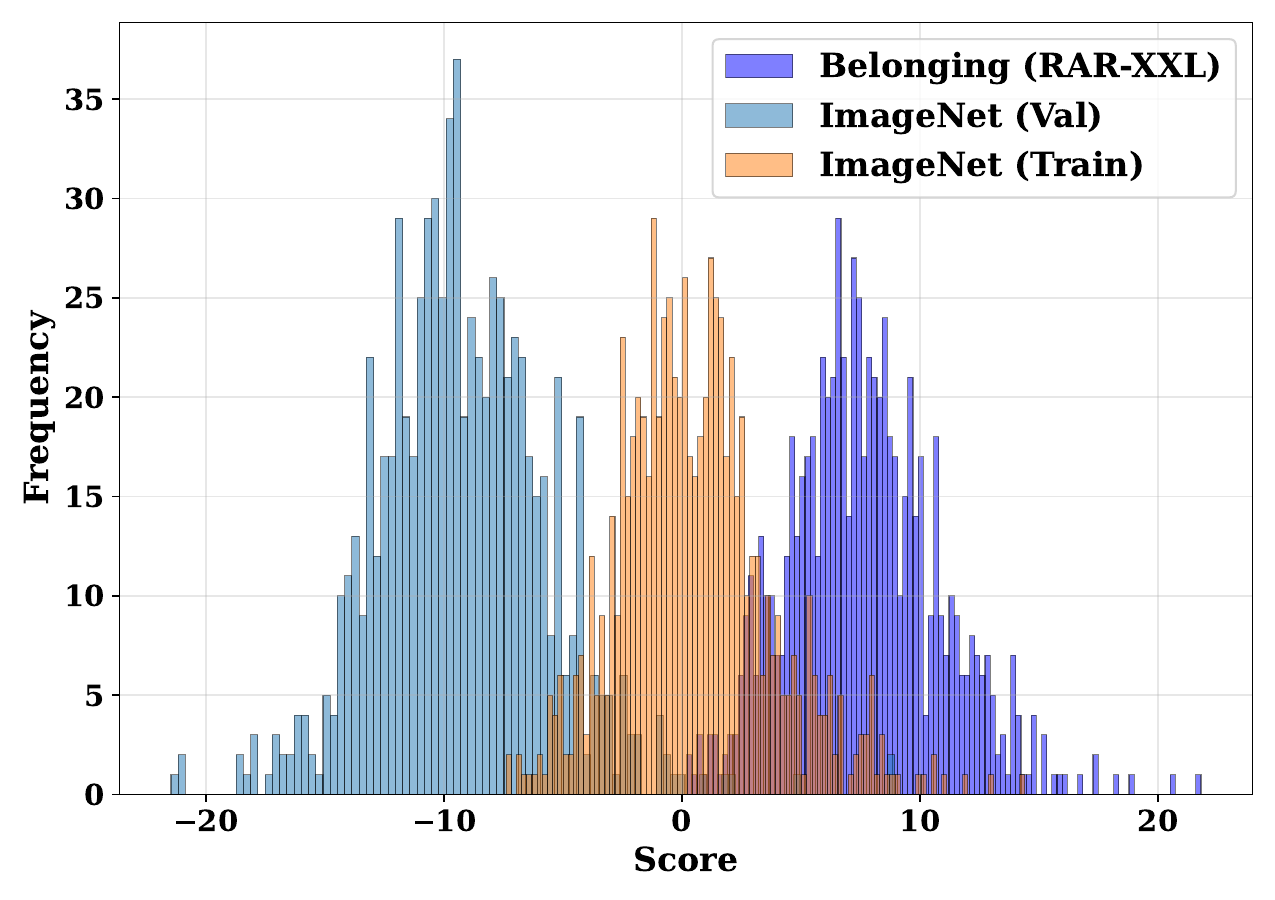}
        \caption{Distribution of scores for a IAR-generated image attribution PRADA~\cite{damm2025prada}.}
        \label{fig:prada_fpr_rar}  %
    \end{subfigure}
    \caption{\textbf{Distributions of scores for membership inference and image attribution on RAR-XXL~\cite{yu2025randomized}.}
    }
    \label{fig:delta_prada_fpr_rar}
\end{figure*}

\textbf{RAR.}
\Cref{fig:delta_prada_fpr_rar} shows the score distributions for PIAR~\cite{kowalczuk2025privacyIARs} and the PRADA image attribution method~\cite{damm2025prada} on RAR-XXL.
For both scoring functions, the generated (``Belonging'') distribution is substantially closer to the training distribution than the held-out validation set.
This confirms that the conditional probability discrepancy used by existing MIAs cannot reliably separate members from generated samples in the RAR architecture.

\begin{figure*}[t]
    \centering
    \begin{subfigure}[b]{0.49\textwidth}
        \centering
        \includegraphics[width=\textwidth]{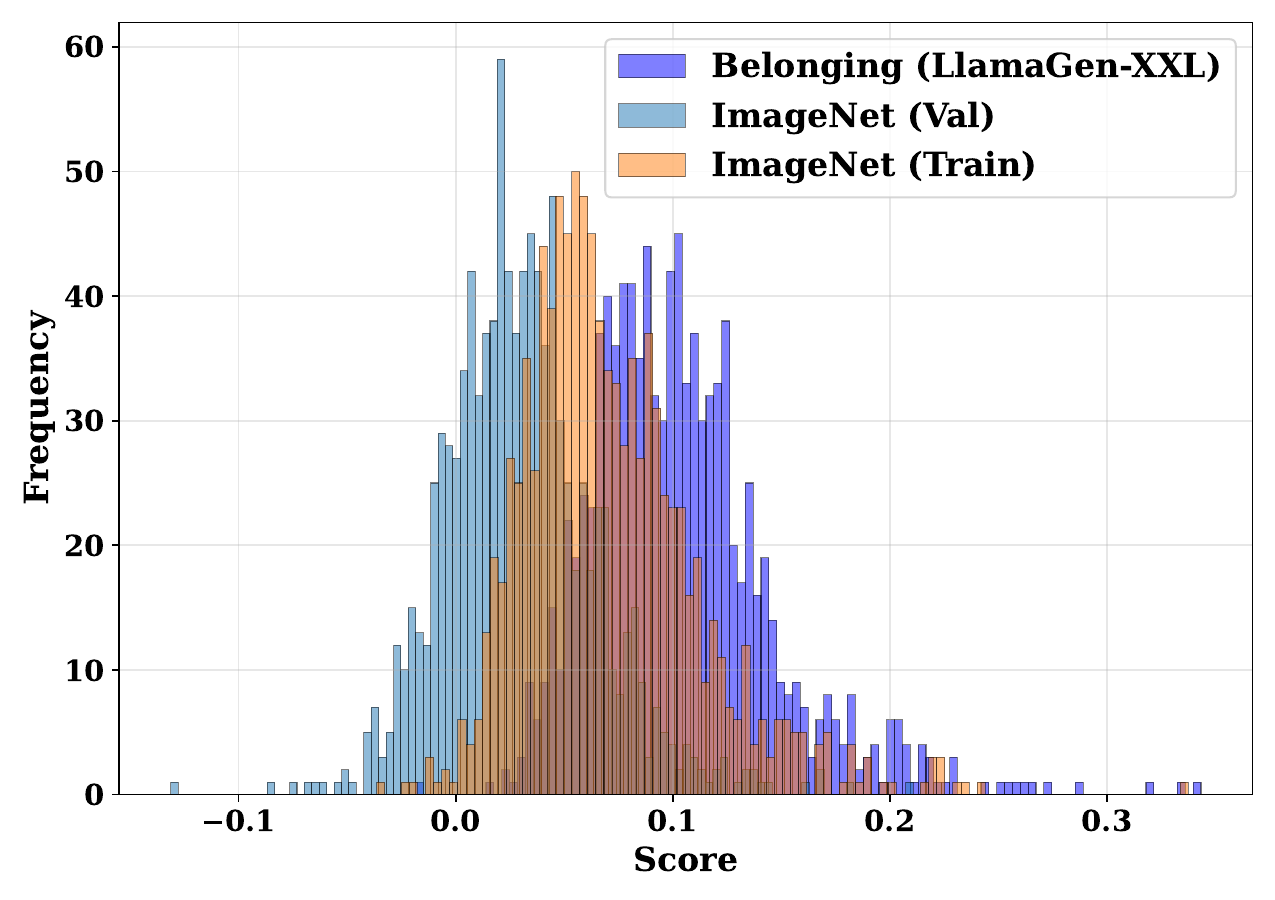}
        \caption{The distribution of scores for the membership inference attack PIAR~\cite{kowalczuk2025privacyIARs}}
        \label{fig:delta_fpr_llamagen}  %
    \end{subfigure}
    \hfill
    \begin{subfigure}[b]{0.49\textwidth}
        \centering
        \includegraphics[width=\textwidth]{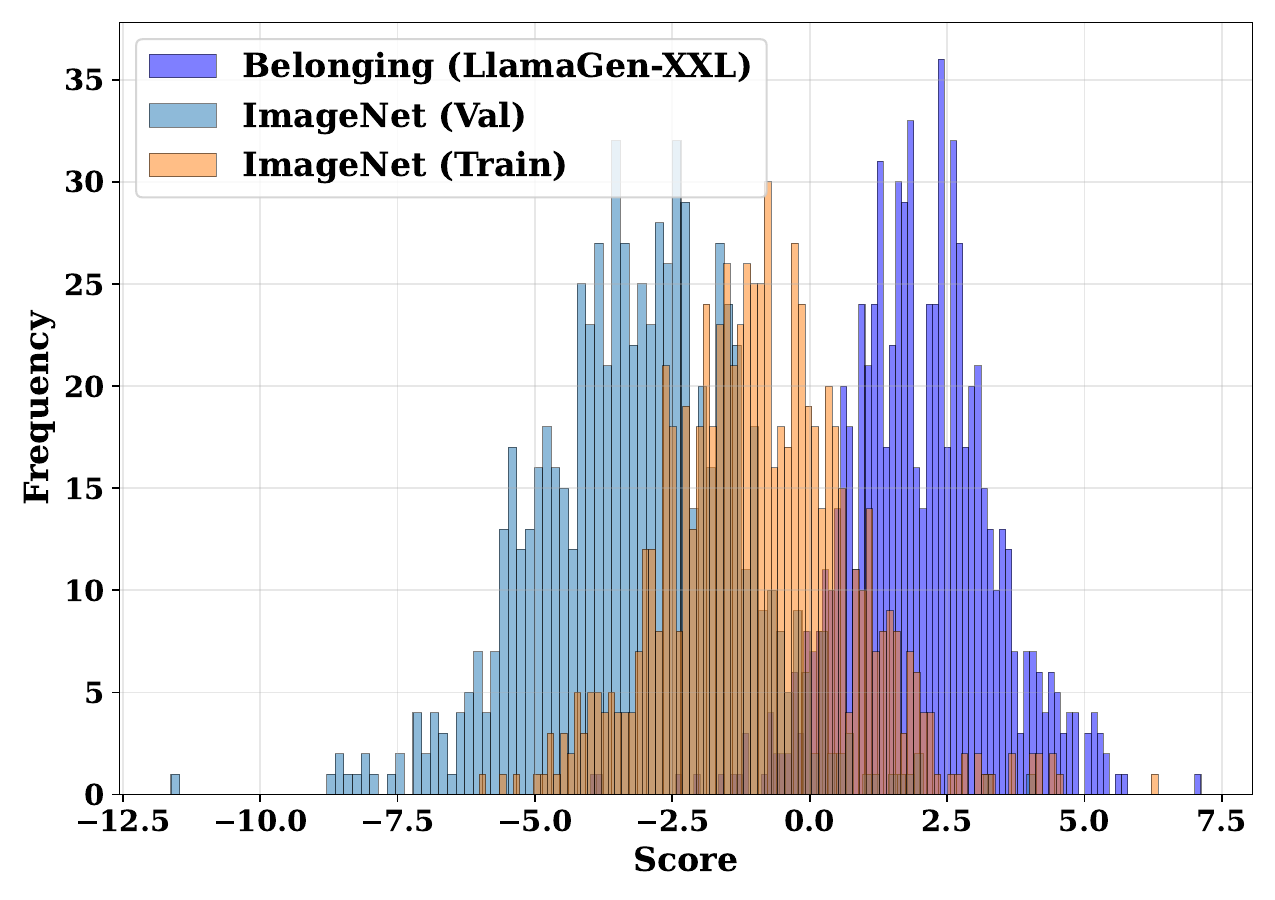}
        \caption{The distribution of scores for a IAR-generated image attribution~\cite{damm2025prada}.}
        \label{fig:prada_fpr_llamagen}  %
    \end{subfigure}
    \caption{\textbf{Distributions of scores for membership inference and image attribution on LlamaGen-XXL~\cite{var_tian2024visualautoregressivemodelingscalable}.} 
    }
    \label{fig:delta_prada_fpr_llamagen}
\end{figure*}

\textbf{LlamaGen.}
A similar pattern emerges for LlamaGen-XXL (\Cref{fig:delta_prada_fpr_llamagen}).
Here, the overlap between the generated and training distributions is even more pronounced under the MIA score, with the generated distribution shifted further toward the member region compared to the validation distribution.
The PRADA attribution score provides somewhat better separation, yet a significant fraction of generated samples still falls within the range of training member scores, highlighting the inadequacy of likelihood-based methods alone for the MGI task.

\begin{figure*}[t!]
    \centering
    \begin{subfigure}[b]{0.49\textwidth}
        \centering
        \includegraphics[width=\textwidth]{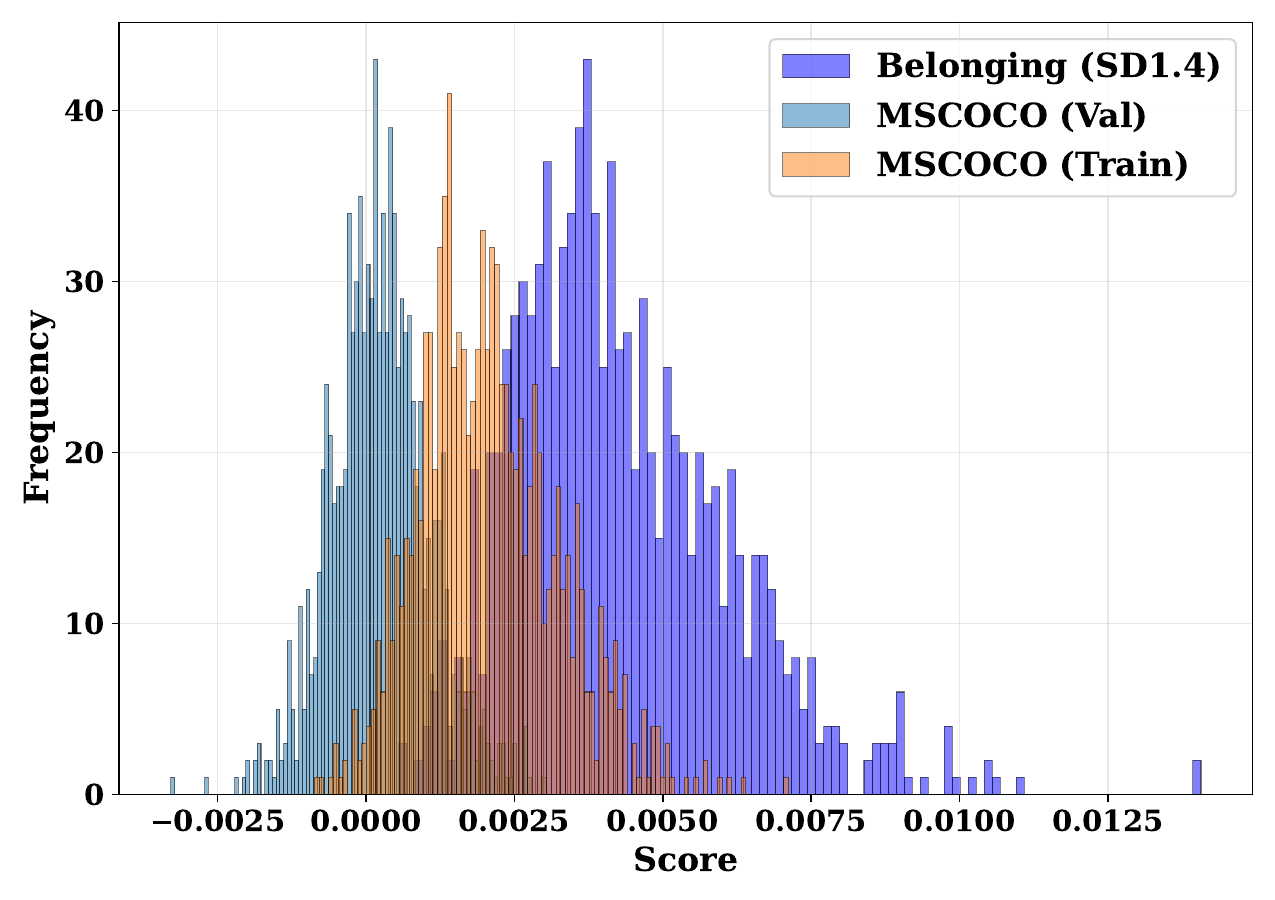}
        \caption{The distribution of scores for the state-of-the-art MIA of Diffusion Models, CLiD~\cite{zhai2024clid}. The evaluated model is $\Mone$.}
        \label{fig:sd14_delta_fpr}  %
    \end{subfigure}
    \hfill
    \begin{subfigure}[b]{0.49\textwidth}
        \centering
        \includegraphics[width=\textwidth]{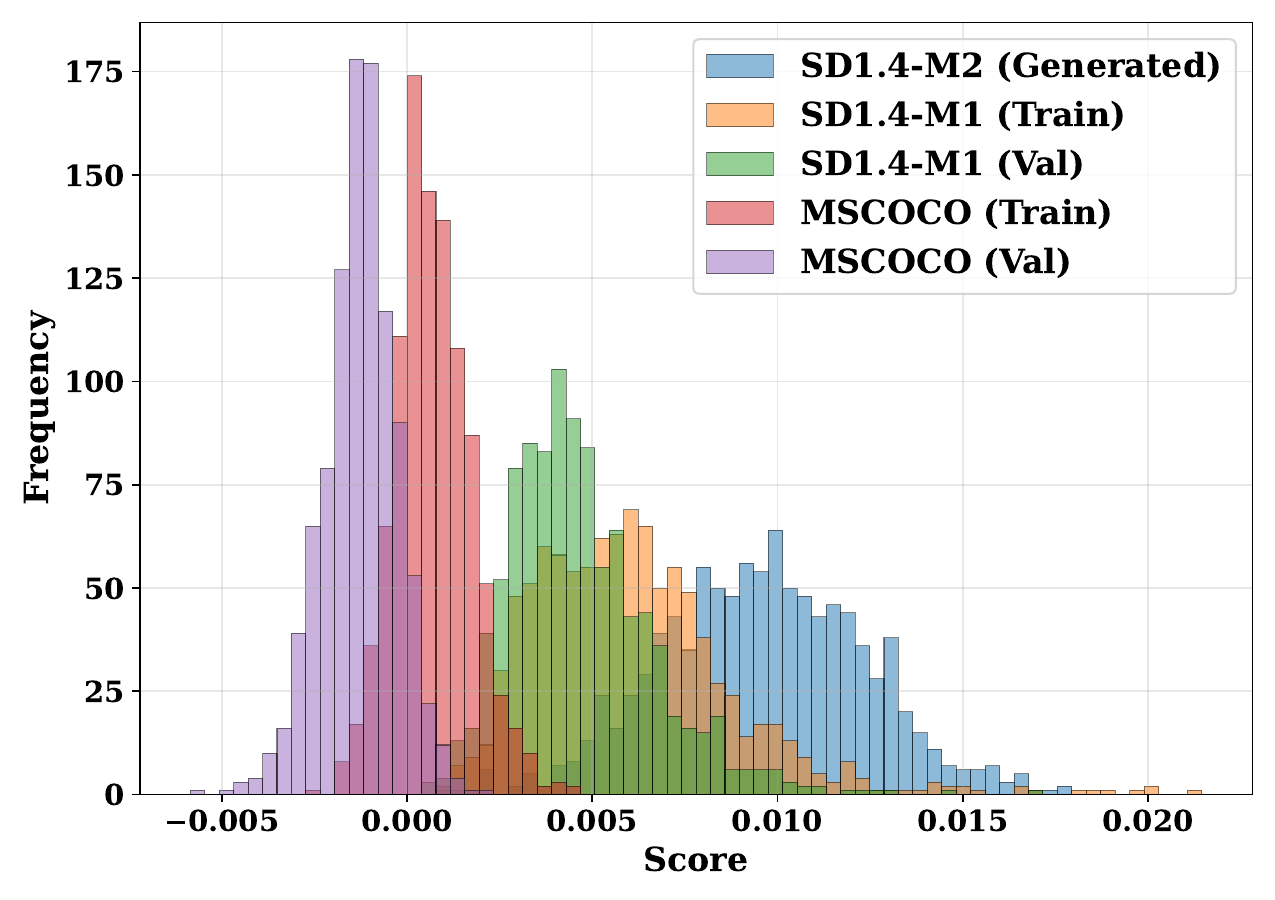}
        \caption{The distribution of scores for the state-of-the-art MIA of Diffusion Models, CLiD~\cite{zhai2024clid}. The evaluated model is $\Mtwo$.}
        \label{fig:sd14_m2_delta_fpr}  %
    \end{subfigure}
    \caption{\textbf{Distributions of scores for membership inference attack on diffusion models for both $\Mone$ and $\Mtwo$.} The evaluated model is Stable Diffusion 1.4 (fine-tuned on MS-COCO).
    }
    \label{fig:sd14_m1_m2_fpr}
\end{figure*}

\textbf{Stable Diffusion~1.4.}
We extend the analysis to diffusion models in \Cref{fig:sd14_m1_m2_fpr}, which plots the CLiD MIA score~\cite{zhai2024clid} distributions for both $\Mone$ and $\Mtwo$.
For $\Mone$, the generated images exhibit score distributions that overlap substantially with training members, consistent with the findings on IARs.
In the $\Mtwo$ setting, the additional fine-tuning on generated data introduces even more complex membership signals, resulting in a more entangled set of distributions.
These results motivate the multi-stage approach of DCB, which leverages complementary autoencoder-based signals to disentangle these overlapping distributions.

\subsection{Model Derivative Setting}

\begin{figure*}[t!]
    \centering
    \begin{subfigure}[b]{0.49\textwidth}
        \centering
        \includegraphics[width=\textwidth]{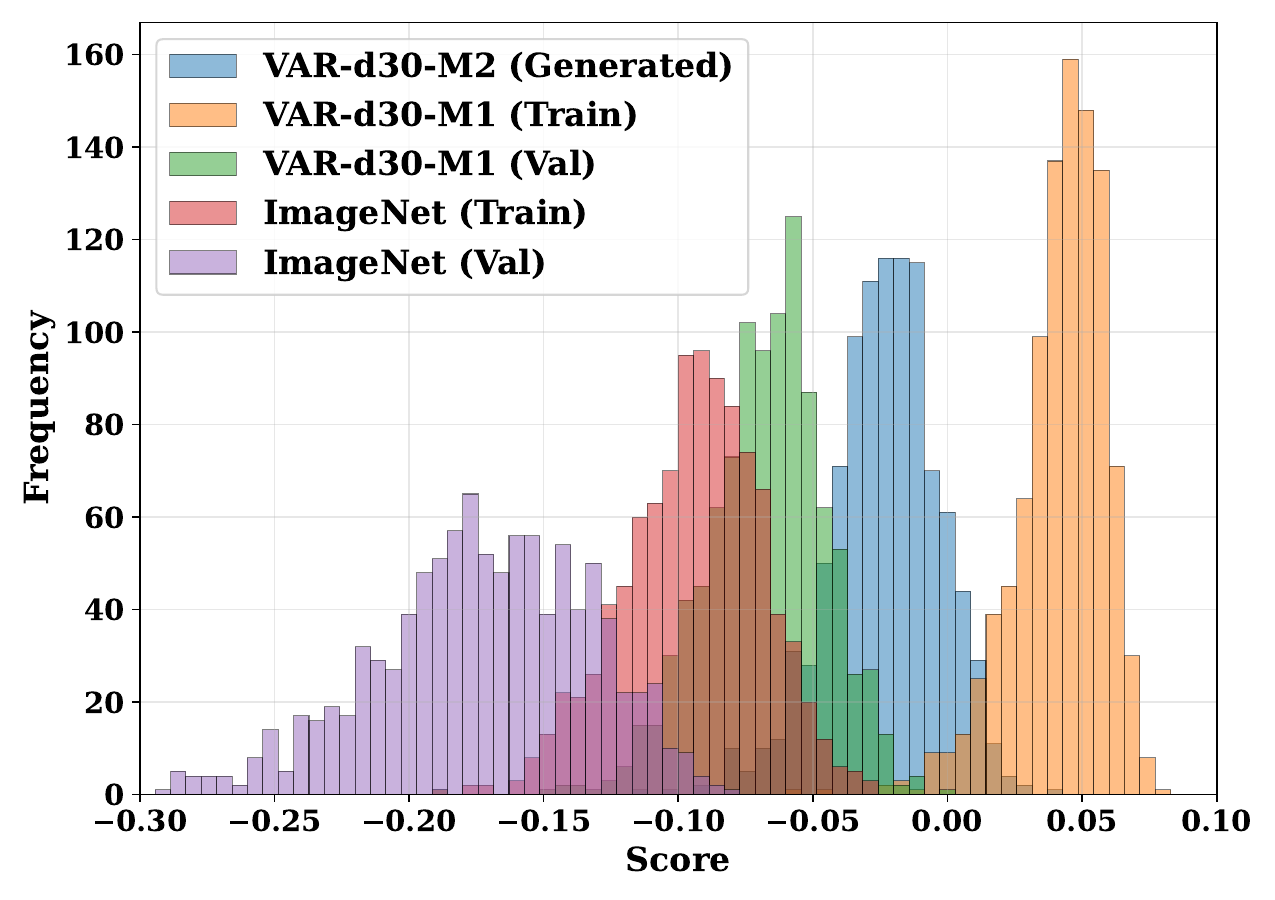}
        \caption{Distribution for VAR-d30~\cite{var_tian2024visualautoregressivemodelingscalable}.}
        \label{fig:m2_mia_var}  %
    \end{subfigure}
    \hfill
    \begin{subfigure}[b]{0.49\textwidth}
        \centering
        \includegraphics[width=\textwidth]{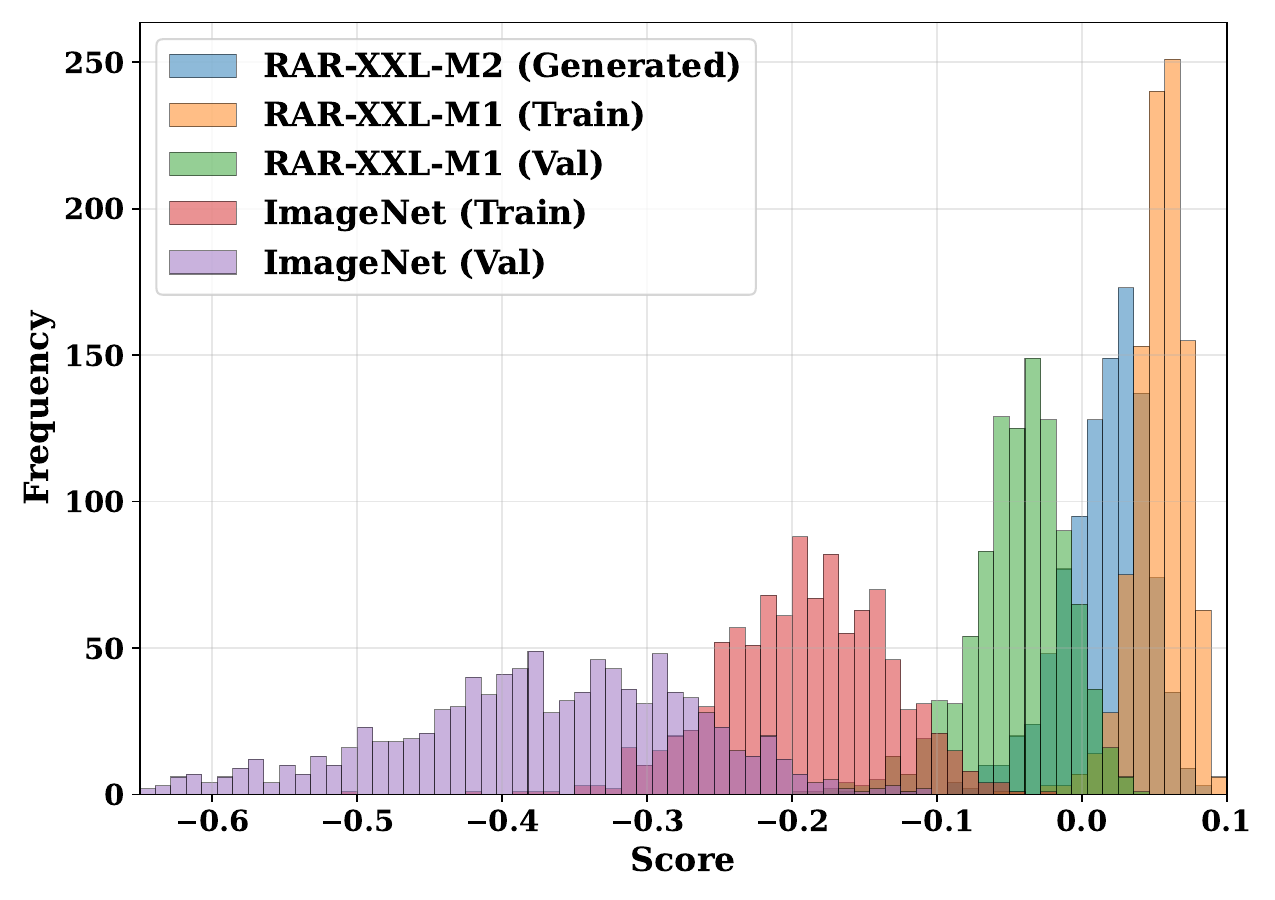}
        \caption{Distribution for RAR-XXL~\cite{yu2025randomized}.}
        \label{fig:m2_mia_rar}  %
    \end{subfigure}
    \caption{\textbf{Distributions of scores for the state-of-the-art MIA on IARs~\cite{yu2025icas} in the model ($\Mtwo$) derivative setting.} With respect to the model $\Mtwo$, we assign the following labels to the datasets: \textit{Generated} for $\Mtwo$-generated data, \textit{Train} for the member data used for training $\Mtwo$ (including pre-training and finetuning), and \textit{Val} for non-member validation data. For the model $\Mtwo$, generated members ($G_M$) and generated non-members ($G_N$) yield high MIA scores that overlap with or exceed those of natural members, so probability-based MIAs alone cannot distinguish the provenance of generated samples.
    }
    \label{fig:m2_mia}
\end{figure*}

\begin{figure*}[t!]
    \centering
    \begin{subfigure}[b]{0.49\textwidth}
        \centering
        \includegraphics[width=\textwidth]{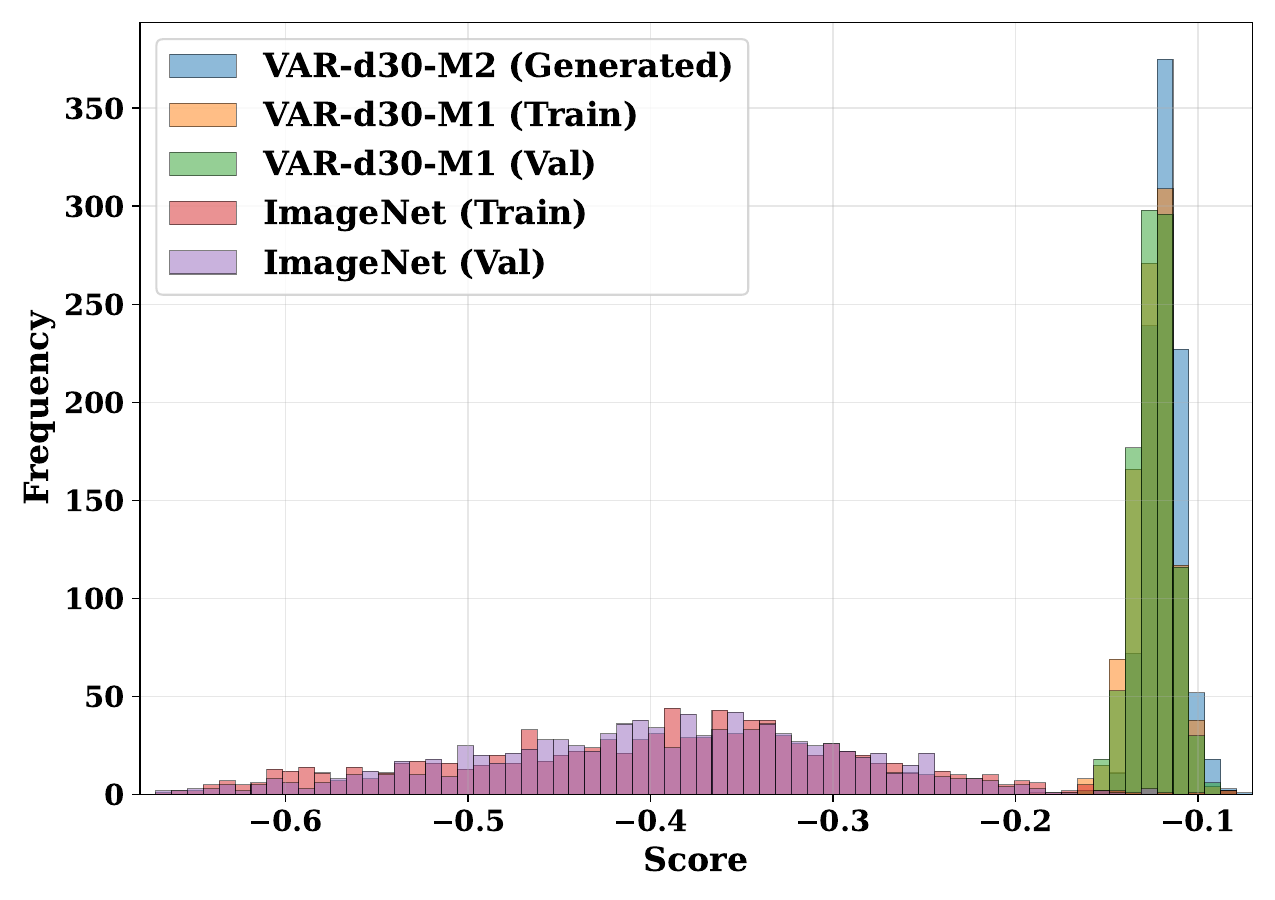}
        \caption{Distribution for VAR-d30~\cite{var_tian2024visualautoregressivemodelingscalable}.}
        \label{fig:m2_ae_var}  %
    \end{subfigure}
    \hfill
    \begin{subfigure}[b]{0.49\textwidth}
        \centering
        \includegraphics[width=\textwidth]{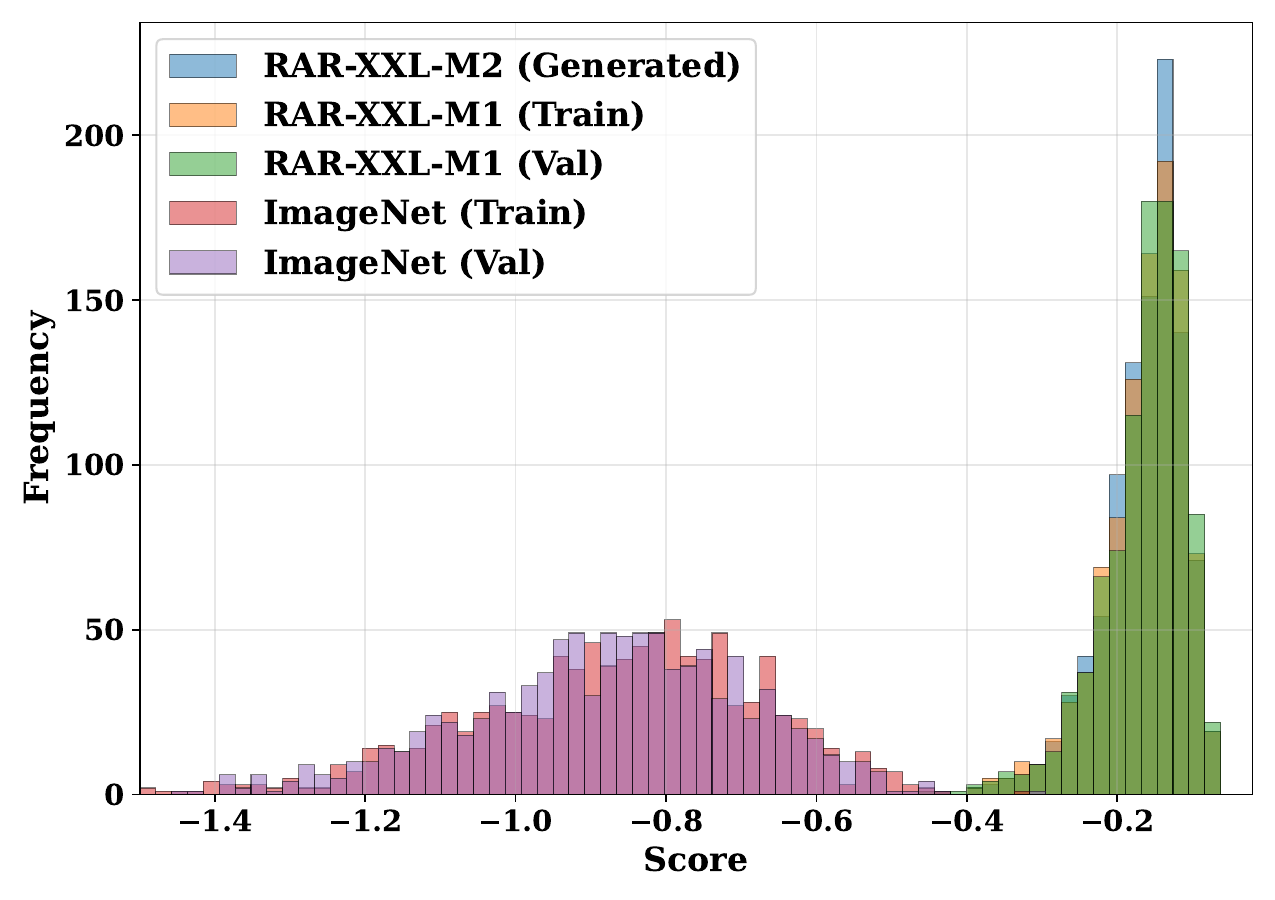}
        \caption{Distribution for RAR-XXL~\cite{yu2025randomized}.}
        \label{fig:m2_ae_rar}  %
    \end{subfigure}
    \caption{\textbf{Distributions of scores for autoencoder-based score in the model ($\Mtwo$) derivative setting.} With respect to the model $\Mtwo$, we assign the following labels to the datasets: \textit{Generated} for $\Mtwo$-generated data, \textit{Train} for the member data used for training $\Mtwo$ (including pre-training and finetuning), and \textit{Val} for non-member validation data. The evaluated autoencoder-based score is defined by~\Cref{eq:ae_score}. The score cleanly separates natural from generated images, as generated images exhibit lower reconstruction and quantization errors, but it cannot distinguish among the different sources of generated content ($G_M$ vs\ $G_N$ vs\ $G'$).
    }
    \label{fig:m2_ae}
\end{figure*}

\begin{figure*}[t!]
    \centering
    \begin{subfigure}[b]{0.49\textwidth}
        \centering
        \includegraphics[width=\textwidth]{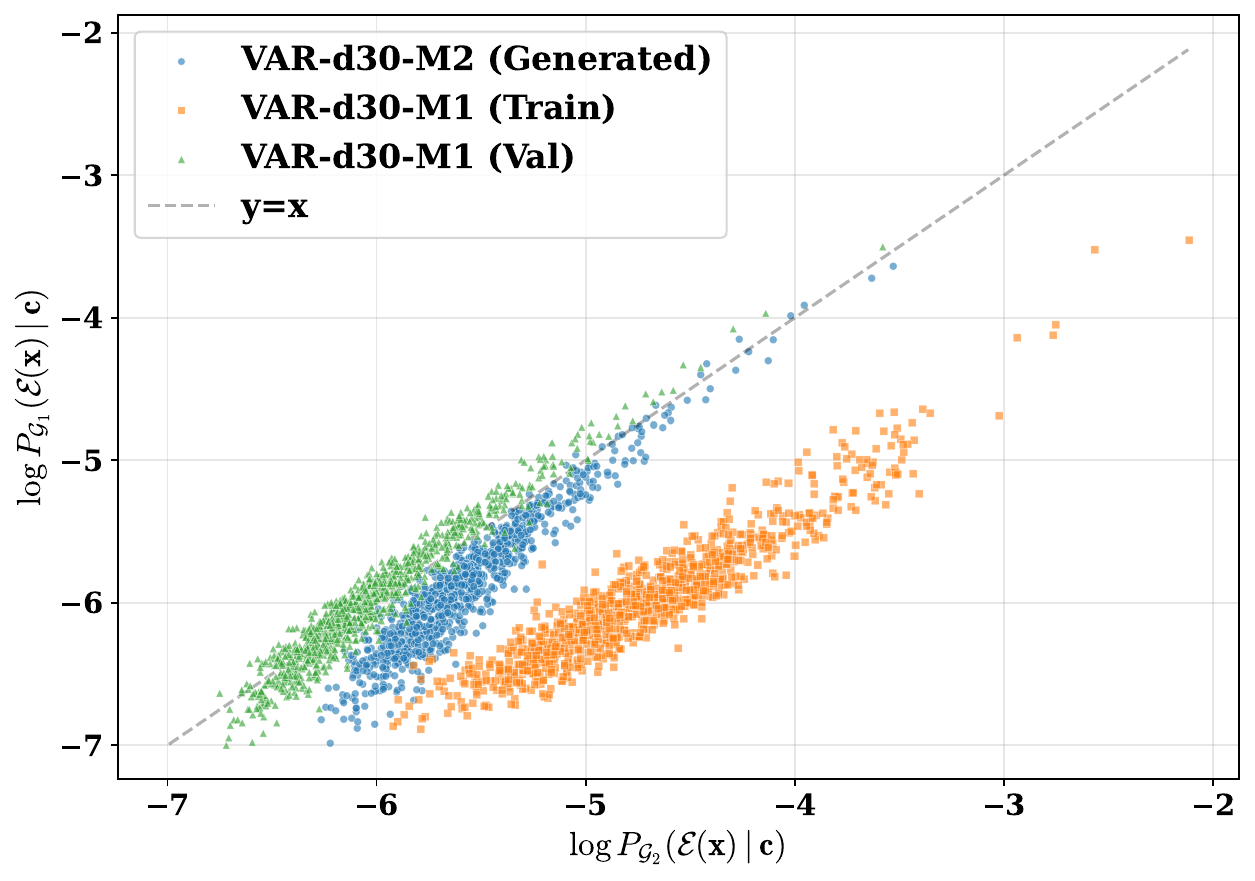}
        \caption{Distribution for VAR-d30~\cite{var_tian2024visualautoregressivemodelingscalable}.}
        \label{fig:m2_cross_var}  %
    \end{subfigure}
    \hfill
    \begin{subfigure}[b]{0.49\textwidth}
        \centering
        \includegraphics[width=\textwidth]{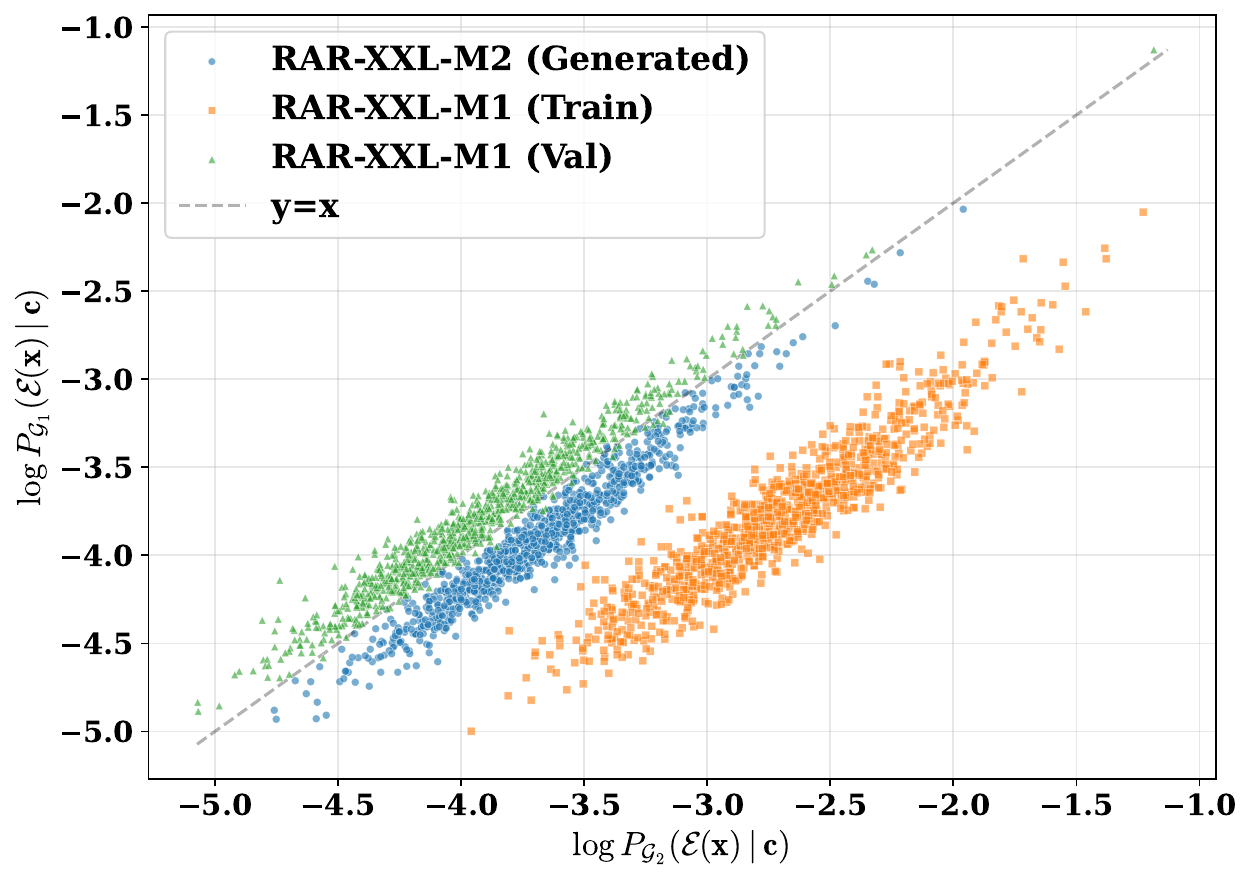}
        \caption{Distribution for RAR-XXL~\cite{yu2025randomized}.}
        \label{fig:m2_cross_rar}  %
    \end{subfigure}
    \caption{\textbf{Distributions of scores for the cross-generator probability discrepancy in the model derivative setting.} With respect to the model $\Mtwo$, we assign the following labels to the datasets: \textit{Generated} for $\Mtwo$-generated data, \textit{Train} for the member data used for training $\Mtwo$ (including pre-training and finetuning), and \textit{Val} for non-member validation data. Comparing the conditional log-probabilities under $\gG_1$ and $\gG_2$ reveals distinct clusters for images generated by $\Mone$ versus $\Mtwo$, enabling fine-grained attribution among generated sources. Together with the MIA and autoencoder signals, these three complementary scoring functions motivate the cascaded design of DCB.
    }
    \label{fig:m2_cross}
\end{figure*}

We additionally visualize the score distributions for the model derivative setting.
\Cref{fig:m2_mia,fig:m2_ae,fig:m2_cross} present three complementary views under the $\Mtwo$ scenario, each shown for both VAR and RAR-XXL: the latent-generator MIA score, the autoencoder reconstruction and quantization error, and the cross-generator probability discrepancy, respectively.

The MIA score distribution in \Cref{fig:m2_mia} reveals a complex mixture: generated members ($G_M$) and generated non-members ($G_N$) produce high MIA scores that overlap with or exceed those of natural members.
This confirms that probability-based MIAs alone cannot distinguish the provenance of generated samples in the derivative setting.

In contrast, the autoencoder-based score in \Cref{fig:m2_ae} provides a clear separation between natural and generated images, as generated images exhibit lower reconstruction and quantization errors.
However, it cannot distinguish among different sources of generated content (\ie, $G_M$ vs.\ $G_N$ vs.\ $G'$).

The cross-generator probability discrepancy in \Cref{fig:m2_cross} addresses this gap: by comparing the conditional log-probabilities under $\gG_1$ and $\gG_2$, the feature vector $\rvphi(\rvx, \rvc)$ reveals distinct clusters for images generated by $\Mone$ versus $\Mtwo$, enabling fine-grained attribution among generated sources.
Together, these three complementary signals form the basis of the DCB pipeline and motivate its cascaded three-stage design.

\section{Additional Results}\label{app:auc_results}

\subsection{Model Direct Training Setting}

\textbf{AUC.}
\Cref{table:m1_auc} and \Cref{table:dm_m1_auc} report the AUC for IARs and DMs, respectively.
DCB achieves an overall AUC of $98.0$ on IARs and $96.5$ on DMs, substantially outperforming all baselines.
Notably, the advantage of DCB is most pronounced in the $N_M/G$ comparison, where existing MIAs exhibit very limited AUC (\eg, $7.8$ for PIAR and $0.9$ for ICAS on RAR), confirming that likelihood-based MIAs cannot separate members from generated samples.
DCB achieves $100.0$ AUC on both RAR and LlamaGen for this critical comparison, demonstrating perfect separation through its autoencoder-based filtering stage.
For DMs, PRADA degrades to below $52$ AUC across both SD~1.4 and SD~2.1, while DCB reaches at least $99.9$.

\textbf{TPR@5\%FPR.}
\Cref{table:m1_tpr5} and \Cref{table:dm_m1_tpr5} present the TPR at 5\% FPR.
The trends are consistent with the AUC results: DCB achieves an overall TPR@5\%FPR of $92.7$ on IARs and $85.4$ on DMs, compared to the best baseline scores of $71.7$ (PRADA on IARs) and $50.5$ (CLiD/ICAS on DMs).
The improvement is especially significant in the $N_M/G$ column, where DCB achieves $100.0\%$ TPR@5\%FPR on RAR and LlamaGen while all baselines remain below $87\%$.
For the $N_M/N_N$ comparison, DCB matches the best baseline in each case, as it falls back on the standard MIA score in Stage~2 of the pipeline after filtering out generated samples.

\ifcvpr
\begin{table*}[t!]
\else
\begin{table}[t!]
\fi
    \centering
    \caption{\textbf{AUC for IARs in the direct training setting.} }
    \label{table:m1_auc}
    \tiny
    \begin{tabular}{lc*{11}{c}c}
        \toprule
        \multirow{2}{*}{Method} & \multicolumn{4}{c}{RAR} & \multicolumn{4}{c}{VAR} & \multicolumn{4}{c}{LlamaGen} & \multirow{2}{*}{Overall} \\
        \cmidrule(lr){2-5} \cmidrule(lr){6-9} \cmidrule(lr){10-13}
        & $N_M$/$G$ & $N_N$/$G$ & $N_M$/$N_N$ & Avg & $N_M$/$G$ & $N_N$/$G$ & $N_M$/$N_N$ & Avg & $N_M$/$G$ & $N_N$/$G$ & $N_M$/$N_N$ & Avg & \\
        \midrule
        PIAR & 7.8 & 99.8 & 98.4 & 68.7 & 96.6 & 95.1 & 99.6 & 97.1 & 28.0 & 92.5 & 79.7 & 66.7 & 77.5 \\
        ICAS & 0.9 & 100.0 & 98.6 & 66.5 & 97.0 & 95.3 & \textbf{99.9} & 97.4 & 6.3 & 99.2 & \textbf{84.3} & 63.3 & 75.7 \\
        PRADA & 98.0 & 100.0 & \textbf{99.1} & 99.0 & 3.4 & 95.9 & 99.8 & 66.4 & 93.1 & 98.9 & 79.1 & 90.4 & 85.3 \\
        Ours & \textbf{100.0} & \textbf{100.0} & 98.6 & \textbf{99.5} & \textbf{99.6} & \textbf{99.8} & \textbf{99.9} & \textbf{99.8} & \textbf{100.0} & \textbf{100.0} & \textbf{84.3} & \textbf{94.8} & \textbf{98.0} \\
        \bottomrule
    \end{tabular}
\ifcvpr
\end{table*}
\else
\end{table}
\fi

\ifcvpr
\begin{table*}[t!]
\else
\begin{table}[t!]
\fi
    \centering
    \caption{\textbf{AUC for the DMs the direct training setting.} }
    \label{table:dm_m1_auc}
    \tiny
    \begin{tabular}{lc*{11}{c}c}
        \toprule
        \multirow{2}{*}{Method} & \multicolumn{4}{c}{SD1.4} & \multicolumn{4}{c}{SD2.1} & \multirow{2}{*}{Overall} \\
        \cmidrule(lr){2-5} \cmidrule(lr){6-9}
        & $N_M$/$G$ & $N_N$/$G$ & $N_M$/$N_N$ & Avg & $N_M$/$G$ & $N_N$/$G$ & $N_M$/$N_N$ & Avg & \\
        \midrule
        CLiD & 14.0 & 99.3 & \textbf{90.8} & 68.1 & 14.7 & 98.8 & \textbf{88.2} & 67.2 & 67.6 \\
        ICAS & 14.0 & 99.3 & 90.8 & 68.1 & 14.7 & 98.8 & 88.2 & 67.2 & 67.6 \\
        PRADA & 51.4 & 41.9 & 39.8 & 44.4 & 53.1 & 44.3 & 40.7 & 46.0 & 45.2 \\
        Ours & \textbf{99.9} & \textbf{100.0} & 90.8 & \textbf{96.9} & \textbf{100.0} & \textbf{100.0} & 88.2 & \textbf{96.1} & \textbf{96.5} \\
        \bottomrule
    \end{tabular}
\ifcvpr
\end{table*}
\else
\end{table}
\fi

\ifcvpr
\begin{table*}[t!]
\else
\begin{table}[t!]
\fi
    \centering
    \caption{\textbf{TPR@5\%FPR for IARs in the direct training setting.} }
    \label{table:m1_tpr5}
    \tiny
    \begin{tabular}{lc*{11}{c}c}
        \toprule
        \multirow{2}{*}{Method} & \multicolumn{4}{c}{RAR} & \multicolumn{4}{c}{VAR} & \multicolumn{4}{c}{LlamaGen} & \multirow{2}{*}{Overall} \\
        \cmidrule(lr){2-5} \cmidrule(lr){6-9} \cmidrule(lr){10-13}
        & $N_M$/$G$ & $N_N$/$G$ & $N_M$/$N_N$ & Avg & $N_M$/$G$ & $N_N$/$G$ & $N_M$/$N_N$ & Avg & $N_M$/$G$ & $N_N$/$G$ & $N_M$/$N_N$ & Avg & \\
        \midrule
        PIAR & 0.3 & \textbf{100.0} & 94.1 & 64.8 & 82.2 & 64.8 & 99.7 & 82.2 & 2.2 & 59.2 & 30.3 & 30.6 & 59.2 \\
        ICAS & 0.0 & 99.9 & 93.3 & 64.4 & 87.1 & 77.3 & 99.9 & 88.1 & 0.0 & 96.2 & \textbf{41.8} & 46.0 & 66.2 \\
        PRADA & 86.5 & \textbf{100.0} & \textbf{97.5} & 94.7 & 0.0 & 82.7 & \textbf{100.0} & 60.9 & 52.5 & 97.2 & 28.8 & 59.5 & 71.7 \\
        Ours & \textbf{100.0} & \textbf{100.0} & 93.3 & \textbf{97.8} & \textbf{99.4} & \textbf{99.6} & 99.9 & \textbf{99.6} & \textbf{100.0} & \textbf{100.0} & \textbf{41.8} & \textbf{80.6} & \textbf{92.7} \\
        \bottomrule
    \end{tabular}
\ifcvpr
\end{table*}
\else
\end{table}
\fi

\ifcvpr
\begin{table*}[t!]
\else
\begin{table}[t!]
\fi
    \centering
    \caption{\textbf{TPR@5\%FPR for DMs in the direct training setting.} }
    \label{table:dm_m1_tpr5}
    \tiny
    \begin{tabular}{lc*{11}{c}c}
        \toprule
        \multirow{2}{*}{Method} & \multicolumn{4}{c}{SD1.4} & \multicolumn{4}{c}{SD2.1} & \multirow{2}{*}{Overall} \\
        \cmidrule(lr){2-5} \cmidrule(lr){6-9}
        & $N_M$/$G$ & $N_N$/$G$ & $N_M$/$N_N$ & Avg & $N_M$/$G$ & $N_N$/$G$ & $N_M$/$N_N$ & Avg & \\
        \midrule
        CLiD & 0.0 & 96.4 & \textbf{58.5} & 51.6 & 0.0 & 94.2 & \textbf{54.2} & 49.5 & 50.5 \\
        ICAS & 0.0 & 96.4 & \textbf{58.5} & 51.6 & 0.0 & 94.2 & \textbf{54.2} & 49.5 & 50.5 \\
        PRADA & 7.3 & 2.9 & 2.3 & 4.2 & 4.0 & 1.9 & 2.1 & 2.7 & 3.4 \\
        Ours & \textbf{99.9} & \textbf{100.0} & \textbf{58.5} & \textbf{86.1} & \textbf{100.0} & \textbf{100.0} & \textbf{54.2} & \textbf{84.7} & \textbf{85.4} \\
        \bottomrule
    \end{tabular}
\ifcvpr
\end{table*}
\else
\end{table}
\fi

\ifcvpr
\begin{table*}[t!]
\else
\begin{table}[t!]
\fi
    \centering
    \caption{\textbf{AUC for the model derivative setting.}}
    \label{table:m2_auc}
    \tiny
    \resizebox{\textwidth}{!}{
    \begin{tabular}{llc*{9}{c}c}
        \toprule
        \multirow{2}{*}{Model} & \multirow{2}{*}{Method} & \multicolumn{6}{c}{Natural v.s. Generated} & \multicolumn{3}{c}{Among Generated}  & \multicolumn{1}{c}{Natural} & \multirow{2}{*}{Overall} \\
        \cmidrule(lr){3-8}\cmidrule(lr){9-11}\cmidrule(lr){12-12}
        & & $N_M$/$G_M$ & $N_M$/$G_N$ & $N_M$/$G'$ & $N_N$/$G_M$ & $N_N$/$G_N$ & $N_N$/$G'$
        & $G_M$/$G_N$ & $G_M$/$G'$ & $G_N$/$G'$
        & $N_M$/$N_N$
        & \\
        \midrule
        \multirow{4}{*}{VAR}
        & PIAR & 98.4 & 38.9 & 74.4 & 100.0 & 99.0 & 99.8 & 99.4 & 96.2 & 86.6 & 99.2 & 89.2 \\
        & ICAS & \textbf{100.0} & 79.1 & 98.6 & \textbf{100.0} & 99.4 & \textbf{100.0} & 99.9 & 99.0 & 93.3 & 99.2 & 96.9 \\
        & PRADA & 99.3 & 33.9 & 80.6 & 100.0 & 99.1 & 99.9 & 99.6 & 2.9 & 86.6 & \textbf{99.4} & 80.1 \\
        \cdashline{2-13}
        \addlinespace[2pt]
        & Ours & 99.6 & \textbf{99.6} & \textbf{99.7} & 99.8 & \textbf{99.8} & 99.9 & \textbf{100.0} & \textbf{99.9} & \textbf{97.5} & 99.2 & \textbf{99.5} \\
        \midrule
        \multirow{4}{*}{RAR}
        & PIAR & 100.0 & 97.5 & 99.8 & \textbf{100.0} & 100.0 & 100.0 & 98.5 & 87.7 & 86.5 & 98.4 & 96.8 \\
        & ICAS & \textbf{100.0} & 98.7 & 99.9 & \textbf{100.0} & 100.0 & \textbf{100.0} & 99.6 & 88.6 & 91.6 & 98.6 & 97.7 \\
        & PRADA & 100.0 & 96.9 & 99.8 & \textbf{100.0} & 99.9 & 100.0 & 99.0 & 9.8 & 87.2 & \textbf{98.9} & 89.2 \\
        \cdashline{2-13}
        \addlinespace[2pt]
        & Ours & 100.0 & \textbf{100.0} & \textbf{100.0} & 100.0 & \textbf{100.0} & 100.0 & \textbf{100.0} & \textbf{100.0} & \textbf{99.5} & 98.6 & \textbf{99.8} \\
        \midrule
        \multirow{4}{*}{SD1.4}
        & CLiD & 99.2 & 98.3 & 99.9 & 100.0 & 100.0 & 100.0 & 66.4 & 16.4 & 92.5 & \textbf{90.8} & 86.4 \\
        & ICAS & 99.2 & 98.3 & \textbf{99.9} & 100.0 & 100.0 & \textbf{100.0} & 66.3 & 16.4 & 92.5 & 90.8 & 86.3 \\
        & PRADA & 37.4 & 52.3 & 36.2 & 30.3 & 42.5 & 29.9 & 45.3 & 48.2 & 43.7 & 39.6 & 40.5 \\
        \cdashline{2-13}
        \addlinespace[2pt]
        & Ours & \textbf{99.9} & \textbf{99.9} & 99.9 & \textbf{100.0} & \textbf{100.0} & 99.9 & \textbf{96.6} & \textbf{91.6} & \textbf{99.6} & 90.8 & \textbf{97.8} \\
        \midrule
        \multirow{4}{*}{SD2.1}
        & CLiD & 99.3 & 98.2 & \textbf{100.0} & 100.0 & 100.0 & \textbf{100.0} & 66.1 & 12.8 & 94.9 & \textbf{88.2} & 86.0 \\
        & ICAS & 99.3 & 98.2 & 100.0 & 100.0 & 100.0 & \textbf{100.0} & 66.1 & 12.8 & 94.9 & 88.2 & 86.0 \\
        & PRADA & 32.3 & 52.4 & 23.3 & 22.8 & 43.8 & 16.4 & 44.0 & 38.0 & 33.0 & 40.8 & 34.7 \\
        \cdashline{2-13}
        \addlinespace[2pt]
        & Ours & \textbf{100.0} & \textbf{100.0} & 100.0 & \textbf{100.0} & \textbf{100.0} & 100.0 & \textbf{96.2} & \textbf{90.6} & \textbf{99.9} & 88.2 & \textbf{97.5} \\
        \bottomrule
    \end{tabular}
    }
\ifcvpr
\end{table*}
\else
\end{table}
\fi

\subsection{Model Derivative Setting}

\textbf{AUC.}
\Cref{table:m2_auc} reports the AUC across all pairwise comparisons in the model derivative setting.
DCB achieves the highest overall AUC for every model: $99.5$ (VAR), $99.8$ (RAR), $97.8$ (SD~1.4), and $97.5$ (SD~2.1).
The baselines clearly fail on the challenging comparisons \emph{among generated samples} ($G_M/G'$ and $G_N/G'$), where only DCB's cross-generator probability discrepancy (Stage~3) provides reliable separation.
For example, on VAR, PRADA achieves only $2.9$ AUC for $G_M/G'$, while DCB reaches $99.9$.
On the diffusion models, the performance gap is particularly clear in the ``Among Generated'' columns: CLiD/ICAS achieve $16.4$ and $12.8$ AUC for $G_M/G'$ on SD~1.4 and SD~2.1, respectively, whereas DCB attains $91.6$ and $90.6$.

\textbf{TPR@5\%FPR.}
\Cref{table:m2_tpr5} presents the TPR@5\%FPR for the same setting.
The results are consistent with the AUC analysis.
DCB achieves an overall TPR@5\%FPR of $98.3$ (VAR), $99.1$ (RAR), $91.3$ (SD~1.4), and $90.2$ (SD~2.1), outperforming the strongest baselines by margins ranging from $0.9$ (RAR, vs.\ ICAS at $90.4$) to $25.6$ (VAR, vs.\ ICAS at $88.7$) percentage points.
The ``Among Generated'' comparisons exhibit the largest gaps: for $G_M/G'$ on RAR, PRADA achieves $0.1\%$ while DCB reaches $100.0\%$; for $G_N/G'$ on SD~1.4, PIAR/ICAS achieve $68.4\%$ with DCB achieving $100.0\%$.
These results confirm that the multi-stage architecture of DCB is essential for resolving the fine-grained attribution challenges posed by the model derivative setting.

\ifcvpr
\begin{table*}[t!]
\else
\begin{table}[t!]
\fi
    \centering
    \caption{\textbf{TPR@5\%FPR for the model derivative setting.}}
    \label{table:m2_tpr5}
    \tiny
    \resizebox{\textwidth}{!}{
    \begin{tabular}{llc*{9}{c}c}
    \toprule
        \multirow{2}{*}{Model} & \multirow{2}{*}{Method} & \multicolumn{6}{c}{Natural v.s. Generated} & \multicolumn{3}{c}{Among Generated}  & \multicolumn{1}{c}{Natural} & \multirow{2}{*}{Overall} \\
        \cmidrule(lr){3-8}\cmidrule(lr){9-11}\cmidrule(lr){12-12}
        & & $N_M$/$G_M$ & $N_M$/$G_N$ & $N_M$/$G'$ & $N_N$/$G_M$ & $N_N$/$G_N$ & $N_N$/$G'$
        & $G_M$/$G_N$ & $G_M$/$G'$ & $G_N$/$G'$
        & $N_M$/$N_N$
        & \\
        \midrule
        \multirow{4}{*}{VAR}
        & PIAR & 91.8 & 0.8 & 13.7 & \textbf{100.0} & 98.7 & \textbf{100.0} & 98.1 & 83.9 & 34.6 & \textbf{98.7} & 72.0 \\
        & ICAS & \textbf{100.0} & 32.3 & 94.3 & \textbf{100.0} & 96.9 & 99.9 & 99.8 & 96.1 & 70.7 & 97.1 & 88.7 \\
        & PRADA & 98.7 & 0.9 & 33.5 & \textbf{100.0} & 97.1 & \textbf{100.0} & 99.2 & 0.0 & 49.5 & 97.6 & 67.6 \\
        \cdashline{2-13}
        \addlinespace[2pt]
        & Ours & 99.3 & \textbf{99.4} & \textbf{99.6} & 99.6 & \textbf{99.6} & 99.6 & \textbf{100.0} & \textbf{99.4} & \textbf{89.8} & 97.1 & \textbf{98.3} \\
        \midrule
        \multirow{4}{*}{RAR}
        & PIAR & \textbf{100.0} & 88.0 & 99.3 & \textbf{100.0} & \textbf{100.0} & \textbf{100.0} & 91.7 & 48.7 & 38.3 & 94.1 & 86.0 \\
        & ICAS & \textbf{100.0} & 94.1 & 99.8 & \textbf{100.0} & \textbf{100.0} & \textbf{100.0} & 98.3 & 48.9 & 69.2 & 93.3 & 90.4 \\
        & PRADA & \textbf{100.0} & 76.9 & 99.6 & \textbf{100.0} & \textbf{100.0} & \textbf{100.0} & 96.0 & 0.1 & 50.7 & \textbf{95.9} & 81.9 \\
        \cdashline{2-13}
        \addlinespace[2pt]
        & Ours & \textbf{100.0} & \textbf{100.0} & \textbf{100.0} & \textbf{100.0} & \textbf{100.0} & \textbf{100.0} & \textbf{100.0} & \textbf{100.0} & \textbf{97.8} & 93.3 & \textbf{99.1} \\
        \midrule
        \multirow{4}{*}{SD1.4}
        & CLiD & 96.5 & 91.7 & 99.1 & \textbf{100.0} & \textbf{100.0} & \textbf{100.0} & 15.7 & 1.8 & 68.4 & \textbf{58.5} & 73.2 \\
        & ICAS & 96.4 & 91.7 & 99.1 & \textbf{100.0} & \textbf{100.0} & \textbf{100.0} & 15.7 & 1.8 & 68.4 & \textbf{58.5} & 73.2 \\
        & PRADA & 1.9 & 4.9 & 4.8 & 0.9 & 2.3 & 3.7 & 5.1 & 7.7 & 8.0 & 2.0 & 4.1 \\
        \cdashline{2-13}
        \addlinespace[2pt]
        & Ours & \textbf{99.9} & \textbf{99.9} & \textbf{99.3} & \textbf{100.0} & \textbf{100.0} & 99.5 & \textbf{85.2} & \textbf{70.4} & \textbf{100.0} & \textbf{58.5} & \textbf{91.3} \\
        \midrule
        \multirow{4}{*}{SD2.1}
        & CLiD & 97.2 & 91.2 & 99.7 & \textbf{100.0} & \textbf{100.0} & \textbf{100.0} & 18.4 & 0.3 & 75.8 & \textbf{54.2} & 73.7 \\
        & ICAS & 97.3 & 91.2 & 99.7 & \textbf{100.0} & \textbf{100.0} & \textbf{100.0} & 18.5 & 0.3 & 75.7 & \textbf{54.2} & 73.7 \\
        & PRADA & 1.3 & 4.3 & 2.1 & 0.4 & 2.1 & 0.8 & 4.3 & 6.1 & 5.2 & 2.4 & 2.9 \\
        \cdashline{2-13}
        \addlinespace[2pt]
        & Ours & \textbf{100.0} & \textbf{100.0} & \textbf{99.8} & \textbf{100.0} & \textbf{100.0} & 99.9 & \textbf{80.4} & \textbf{68.4} & \textbf{99.8} & \textbf{54.2} & \textbf{90.2} \\
        \bottomrule
    \end{tabular}
    }
\ifcvpr
\end{table*}
\else
\end{table}
\fi

\setlength{\tabcolsep}{3pt} %
\begin{table*}[t]
    \small
    \begin{center}
    \caption{\textbf{Robustness of Stage 1 ($\quantloss$).} We show \tpr on RAR.}
    \label{table:robust}
    \begin{tabular}{lccccc}
    \toprule
    \multirow{2}{*}{Method} & \multicolumn{5}{c}{Attacks} \\
    \cmidrule{2-6}
    & Orig. & JPEG (60) & Resize (0.5) & Saturation (2.0) & Adv. ($\varepsilon$ = 1) \\
    \midrule
    Ours (w/o Aug) & 100.0 & 91.7 & 88.5 & 97.4 & 68.7 \\
    Ours (w/\ \ \  Aug) & 99.6 & 96.1 & 98.4 & 99.2 & 97.4 \\
    \bottomrule
\end{tabular}
    \end{center}
\end{table*}

\section{Results on More Models}
We further evaluate our approach and the baselines on the direct training setting for two SoTA diffusion models, REPA~\cite{yu2025repa} and Lightning DiT~\cite{yao2025vavae}.
\Cref{table:repa_ldit_m1_auc} shows that our approach generalizes effectively to the two SoTA diffusion models, outperforming the baselines. 

\begin{table*}[t]
\centering
\small
\caption{\textbf{DCB on State-of-the-art Diffusion Models, REPA and Lightning DiT.} We report the AUC of the direct training setting.}
\label{table:repa_ldit_m1_auc}
\begin{tabular}{l ccc ccc ccc}
\toprule
\multirow{2}{*}{Model}& \multicolumn{3}{c}{$N_M$/$N_N$} & \multicolumn{3}{c}{$N_M$/$G$} & \multicolumn{3}{c}{$N_N$/$G$} \\
\cmidrule(lr){2-4} \cmidrule(lr){5-7} \cmidrule(lr){8-10}
 & ICAS & PRADA & DCB & ICAS & PRADA & DCB & ICAS & PRADA & DCB \\
\midrule
\textbf{REPA}-SiT-XL/2  & 74.3 & 55.3 & \textbf{74.3} & 54.2 & 57.3 & \textbf{98.4} & 28.5 & 52.3 & \textbf{98.3} \\
\textbf{LightningDiT}-XL & 72.4 & 61.9 & \textbf{72.4} & 62.3 & 67.4 & \textbf{100.0} & 38.3 & 56.3 & \textbf{100.0} \\
\bottomrule
\end{tabular}

\end{table*}

\begin{table*}[t]\centering
\caption{\textbf{Cross-architecture Setting.}}
\label{table:crossarch_auc}
\begin{tabular}{l ccc ccc ccc}
\toprule
 \multirow{2}{*}{Model} & \multicolumn{3}{c}{$G_M$/$G_N$} & \multicolumn{3}{c}{$G_M$/$G'$} & \multicolumn{3}{c}{$G_N$/$G'$} \\
\cmidrule(lr){2-4} \cmidrule(lr){5-7} \cmidrule(lr){8-10}
 & ICAS & PRADA & DCB & ICAS & PRADA & DCB & ICAS & PRADA & DCB \\
\midrule
\textbf{REPA}$\to$\textbf{RAR}        & 90.9 & 91.6 & \textbf{90.9} & 76.9 & 26.4 & \textbf{99.0} & 72.9 & 77.5 & \textbf{99.4} \\
\textbf{LDiT}$\to$\textbf{RAR}        & 87.3 & 87.2 & \textbf{87.3} & 74.6 & 26.3 & \textbf{99.7} & 69.0 & 70.3 & \textbf{99.7} \\
\textbf{SD\,1.4}$\to$\textbf{SD\,2.1} & 79.9 & 60.6 & \textbf{79.9} & 16.9 & 63.6 & \textbf{100.0} & 96.4 & 73.6 & \textbf{99.9} \\
\bottomrule
\end{tabular}
\end{table*}

\begin{table}[t]
\footnotesize
\centering
\caption{\textbf{Prompt estimation.} Captions from BLIP2/LLaVA replace ground-truth (GT) prompts in~\Cref{table:m2_tpr}.}
\label{tab:prompt_estimation}
\begin{tabular}{l ccc ccc}
\midrule
\multirow{2}{*}{Model}  & \multicolumn{3}{c}{ICAS} & \multicolumn{3}{c}{DCB} \\
\cmidrule(lr){2-4} \cmidrule(lr){5-7}
& GT & BLIP2 & LLaVA & GT & BLIP2 & LLaVA \\
\midrule
\textbf{SD\,1.4} & 63.5 & 46.2 & 33.6 & \textbf{82.6} & \textbf{81.5} & \textbf{78.7} \\
\textbf{SD\,2.1} & 64.4 & 43.4 & 28.6 & \textbf{83.7} & \textbf{82.8} & \textbf{76.7} \\
\midrule
\end{tabular}
\end{table}

\begin{table}[t]
    \footnotesize
    \begin{center}
    \caption{\textbf{Performance of Stage 1 on LlamaGen with and without the optional finetuning.} The metric is \tpr.}
    \label{table:dec_inv}
\begin{tabular}{lccc}
    \midrule
    Method & ImageNet & LAION & MS-COCO \\
    \midrule
    Ours (\textit{w/o finetuning}) & 99.9 & 99.6 & 99.9 \\
    Ours (\textit{w/\ \ \ finetuning}) & 100.0 & 100.0 & 100.0 \\
    \midrule
\end{tabular}
    \end{center}
\end{table}

\section{Additional Results for Memorized Samples}\label{app:mem_samples}

\begin{figure*}[t!]
    \centering
    \begin{subfigure}[b]{0.49\textwidth}
        \centering
        \includegraphics[width=0.85\textwidth]{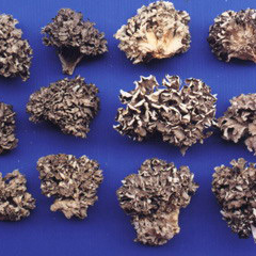}
        \caption{The real training image for the memorized sample.}
        \label{fig:mem_train}  %
    \end{subfigure}
    \hfill
    \begin{subfigure}[b]{0.49\textwidth}
        \centering
        \includegraphics[width=0.85\textwidth]{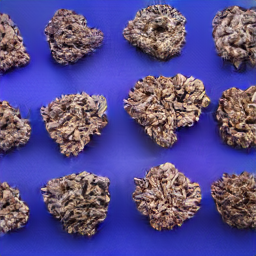}
        \caption{The re-generated image for the memorized sample.}
        \label{fig:mem_regen}  %
    \end{subfigure}
    \caption{\textbf{Visualization of the real training sample and re-generated images for one memorized sample.} The evaluated model RAR-XXL and the SSCD score is 0.827. The ImageNet label for the image pair is 996.
    }
    \label{fig:mem_rar_vis}
\end{figure*}

We provide additional visualizations for the memorization case study discussed in \Cref{subsec:mem_case_study} of the main paper.
\Cref{fig:mem_rar_vis} shows a representative example of a memorized training sample from RAR-XXL alongside its corresponding re-generated output.
Despite sharing nearly identical visual content---with an SSCD similarity score of $0.827$, well above the $0.7$ threshold used to identify memorized samples---the two images are not pixel-identical.
The re-generated image has passed through the full generation pipeline (autoregressive token sampling and decoding), which introduces subtle generation-specific artifacts.
These artifacts are imperceptible to the human eye but are reliably captured by our autoencoder-based attribution score $\mathcal{L}_A$, as shown in~\Cref{fig:ae_memorized} of the main paper, where the quantization and reconstruction error distributions for memorized training images and their re-generated counterparts are well-separated.

This example illustrates the most challenging case for data provenance: the generated image is a near-duplicate of the training sample, yet it was produced through the model's generative process rather than directly copied.
Standard MIAs, which rely on the latent generator's probability scores, assign nearly identical scores to both the original and the re-generated version (\Cref{fig:delta_memorized}), since both are mostly consistent with the learned distribution.
In contrast, DCB's autoencoder-based filtering stage detects the generation artifacts introduced by the encode-decode pipeline, enabling reliable separation even in this extreme memorization regime.
The quantitative performance on all $169$ identified memorized samples is reported in \Cref{tab:example} of the main paper, where DCB achieves $97.5$ AUC and $93.5\%$ TPR@5\%FPR, compared to at most $61.8$ AUC and $3.0\%$ TPR@5\%FPR for the best baseline.

\section{Robustness}
We evaluate our proposed Stage 1 (as described in \Cref{sec:protocol}) under real-world web-pipeline degradations (JPEG, resize, saturation) and the strong adaptive adversarial attack that directly optimizes perturbations to maximize $\quantloss$. The results are shown in \Cref{table:robust}. Without any augmentation, Stage 1 already retains $\geq 88.5\%$ \tpr under all natural transforms. Optionally, we apply augmentations to the finetuning process of the encoder, inspired by~\cite{zhao2026provenance} and~\cite{jovanovic2026watermarking}.
The augmented fine-tuning further boosts robustness to $97.4\%$ even under the adaptive adversarial attack.

\section{Cross-architecture Generalization}

In the model derivative setting, we mainly consider the \textit{identical-architecture} setting where $\Mtwo$ is trained on the data generated by $\Mone$ with the same architecture as $\Mtwo$.
In this section, we further evaluate a \textit{cross-architecture }setting, where $\Mtwo$ is trained on images generated by a model with different architecture.
We note that a cross-architecture setting is an \emph{easier} setting for MGI, not more challenging.
If $\Mtwo$ is trained on images generated by another model architecture, the distinct autoencoder architectures \emph{enhance} Stage~1 separation of $G_M$/$G'$ (rather than collapsing it), and $G_M$/$G_N$ reduces to standard MIA. 
In contrast, the identical-architecture setting is a more challenging setting, because $G_M$,$G_N$, and $G'$ are all from the same autoencoder and therefore require Stage 3.
Therefore, we choose the the more challenging identical-architecture setting for evaluation in our main content.
\Cref{table:crossarch_auc} evaluates the cross-architecture setting on the \textbf{SD\,1.4}$\to$\textbf{SD\,2.1} case and two heterogeneous DM-to-IAR pairs. The results show that DCB attains $\geq 99\%$ AUC on $G_M/G'$ for all settings.

\section{Prompt Estimation}

We note that ground-truth prompts can be absent in certain applications.
In such cases, we use BLIP2/LLaVA to generate prompts for a given image.
\Cref{tab:prompt_estimation} compares the performance of our approach
using groundtruth (GT) and BLIP2/LLaVA-generated captions. 
The results show that DCB still achieves high performance.

\section{Model Access}
Our approaches perform the best with the white-box access, where the optional finetuning is enabled.
We also evaluate a gray-box setting, where only the outputs of the autoencoder and latent generator can be observed.
In the gray-box setting, the optional finetuning is not possible and our proposed approach fully operates with the original, un-finetuned encoder.
\Cref{table:dec_inv} shows that our method achieves high performance on LlamaGen without the optional finetuning.
Moreover, we never use the finetuned encoder for the Diffusion Models.
Our method can operate in a gray-box setting for many models, requiring only loss values and generative model outputs, matching the access assumptions of SOTA MIAs (\eg, PIAR, CLiD, ICAS)

\section{Hyperparameter Analysis}

We evaluate two hyperparameters in the KDE test in Stage~1: density threshold $\alpha$ and bandwidth multiplier $\sigma$. 
\Cref{table:kde_sens}~shows that our approach is not sensitive to these two hyperparameters.

\begin{table}[t]
\caption{\textbf{Stage-1 KDE sensitivity.}}
\vspace{-1em}
\label{table:kde_sens}
\begin{subtable}{0.48\columnwidth}
\centering
\caption{$\alpha$ sweep ($\sigma{=}0.03$)}
\label{tab:kde_alpha}
\begin{tabular}{lccc}
\toprule
 $\alpha$ & 0.03  & 0.05 & 0.07 \\
\midrule
VAR-d30 & 99.5 & 98.4 & 97.2 \\
SD\,2.1 & 98.4 & 97.7 & 96.9 \\
\bottomrule
\end{tabular}
\end{subtable}\hfill
\begin{subtable}{0.48\columnwidth}\centering
\caption{$\sigma$ sweep ($\alpha{=}0.05$)}
\label{tab:kde_sigma}
\begin{tabular}{lccc}
\toprule
 $\sigma$ & 0.10 & 0.30 & 0.50 \\
\midrule
VAR-d30 & 97.4 & 98.4 & 99.1 \\
SD\,2.1 & 97.1 & 97.7 & 98.3 \\
\bottomrule
\end{tabular}
\end{subtable}
\end{table}

\begin{table}[t]\centering
\caption{\textbf{Extension of \Cref{table:m1_tpr} to 5k samples.}}
\label{table:rar_5k_t1}
\scriptsize
\begin{tabular}{lcccc}
\midrule
Method & $N_M$/$G$ & $N_N$/$G$ & $N_M$/$N_N$ & Avg \\
\midrule
PIAR ($\Delta$) &   0.1 (0.0) & 99.2 (99.5) & 59.1 (62.6) & 52.8 (54.0) \\
ICAS            &   0.0 (0.0) & 99.8 (99.7) & 78.0 (72.5) & 59.3 (57.4) \\
PRADA           &  49.7 (62.7) & 99.9 (100.0) & 69.0 (81.3) & 72.9 (81.3) \\
\textbf{\ours}    & 100.0  (99.9) & 99.9 (99.9) & 78.0 (72.5) & 92.6 (90.8) \\
\midrule
\end{tabular}
\end{table}

\begin{table}[t]\centering\tiny
\caption{\textbf{Per-image cost (sec) and \ours-vs-MIA cost ratio.}}
\label{table:cost}

\resizebox{\columnwidth}{!}{%
\begin{tabular}{lccccc}
\midrule
Model & Stage 1 & Stage 2 (MIA) & Stage 3 & \ours/MIA (M1) & \ours/MIA (M1+M2) \\
\midrule
RAR-XXL  & 0.047 & 0.101 & 0.026 & \textbf{1.46}$\times$ & \textbf{1.71}$\times$ \\
SD\,2.1   & 0.240 & 1.510 & 0.747 & \textbf{1.16}$\times$ & \textbf{1.66}$\times$ \\
\midrule
\end{tabular}}
\vspace{-2.8em}
\end{table}

\section{Sample Size Evaluation}

We extend our experiments from 1K samples (\Cref{table:m1_tpr}) to 5K samples on RAR-XXL and observe same trends, as we show in~\Cref{table:rar_5k_t1}.

\section{Computational Cost}

\ours targets the offline auditing regime shared by all SoTA MIAs (PIAR, CLiD, PRADA), not real-time use. As shown in~\Cref{table:cost}, \ours adds only $0.16\times$--$0.46\times$ ($M_1$) and $0.66\times$--$0.71\times$ ($M_2$) on top of a single MIA inference. Additionally, \ours uses forward passes only (no backpropagation) and can be parallelized across images, so throughput scales linearly with GPUs and million-scale audits are practical.

\end{document}